\documentclass{article}

\usepackage{microtype}
\usepackage{graphicx}
\usepackage{subfigure}
\usepackage{booktabs} % for professional tables

\usepackage{algorithm}
\usepackage{algorithmic}
\usepackage{amsmath}
\usepackage{amssymb}

\usepackage{hyperref}

\usepackage[accepted]{icml2018_preprint}

\icmltitlerunning{Regret Minimization for Partially Observable Deep Reinforcement Learning}

\begin{document}

\twocolumn[
\icmltitle{Regret Minimization for Partially Observable Deep Reinforcement Learning}



\icmlsetsymbol{equal}{*}

\begin{icmlauthorlist}
\icmlauthor{Peter Jin}{ucb}
\icmlauthor{Kurt Keutzer}{ucb}
\icmlauthor{Sergey Levine}{ucb}
\end{icmlauthorlist}

\icmlaffiliation{ucb}{Department of Electrical Engineering and Computer Sciences, University of California, Berkeley}

\icmlcorrespondingauthor{Peter Jin}{phj@eecs.berkeley.edu}

\icmlkeywords{deep reinforcement learning, partial observability, regret}

\vskip 0.3in
]



\printAffiliationsAndNotice{}  % leave blank if no need to mention equal contribution

\begin{abstract}
Deep reinforcement learning algorithms that estimate state and state-action value functions have been shown to be effective in a variety of challenging domains, including learning control strategies from raw image pixels. However, algorithms that estimate state and state-action value functions typically assume a fully observed state and must compensate for partial observations by using finite length observation histories or recurrent networks. In this work, we propose a new deep reinforcement learning algorithm based on counterfactual regret minimization that iteratively updates an approximation to an advantage-like function and is robust to partially observed state. We demonstrate that this new algorithm can substantially outperform strong baseline methods on several partially observed reinforcement learning tasks: learning first-person 3D navigation in Doom and Minecraft, and acting in the presence of partially observed objects in Doom and Pong.

\end{abstract}

\section{Introduction}

Many reinforcement learning problems of practical interest have the property of
partial observability,
where observations of state are generally non-Markovian.
Practical deep reinforcement learning algorithms fall into two broad classes,
neither of which satisfactorily deals with partial observability
despite the prevalance of partial observations in the real world.
One class of algorithms consists of value function-based methods
such as deep Q-learning \citep{dqn,dqn-nature},
which are known to be highly sample efficient
but generally assume a Markovian observation space.
The other class of algorithms consists of Monte Carlo policy gradient methods
such as trust region policy optimization \citep{trpo},
which do not need to assume Markovian observations
but are less sample efficient than value function-based methods.
Some policy gradient methods
such as advantage actor-critic \citep{a3c}
introduce the Markov assumption through a critic or state-dependent baseline
to improve sample efficiency.

There are two common workarounds for the problem of partial observation:
(a) learning policies and value functions on finite length observation
histories,
and (b) learning recurrent policies and recurrent value functions.
Finite length observation histories concatenate the most recent
raw observations into a stack of observations,
and are a simple but effective technique to approximate the appearance of
full observability.
Recurrent functions can potentially incorporate an infinite observation history,
but they can be harder to optimize.
When using either finite length observation histories or recurrent functions,
the same value function-based methods and policy gradient methods are
employed with their respective tradeoffs in partially observable domains.

We are interested in
developing methods that combine the best of both value function-based
methods and policy gradient methods for partial observable domains.
That is, can we develop methods that are sample efficient but are also robust
to partial observation spaces?

Our contribution is a new model-free deep reinforcement learning algorithm
based on the principle of regret minimization which does not require access
to a Markovian state.
Our method learns a policy by estimating an
advantage-like function
which approximates a quantity called the counterfactual regret.
Counterfactual regret is central to the family of
counterfactual regret minimization (CFR) \citep{zinkevich07}
algorithms for solving incomplete information games.
Hence we call our algorithm ``advantage-based regret minimization'' (ARM).

We evaluate our approach on three partially observable visual
reinforcement learning tasks:
first-person 3D navigation in Doom and Minecraft \citep{vizdoom,malmo},
avoiding partially observed objects in Doom,
and playing Pong with partial observation perturbations \citep{ale}.
In our experiments,
we find that our method is more robust and offers substantial improvement
over prior methods on partially observable tasks.

\section{Related Work}

Deep reinforcement learning algorithms have been demonstrated to achieve
excellent results on a range of complex tasks, including playing games
\citep{dqn-nature,oh2016minecraft}
and continuous control \citep{trpo,ddpg,levine_finn16}.
Prior deep reinforcement learning algorithms either learn
state or state-action value functions~\citep{dqn},
learn policies using policy gradients~\citep{trpo},
or perform a combination of the two using actor-critic architectures~\citep{a3c}.
Policy gradient methods typically do not need to assume a Markovian state,
but tend to suffer from poor sample complexity, due to their inability to
use off-policy data.
Methods based on learning Q-functions can use replay buffers to include off-policy data,
accelerating learning \citep{ddpg}.
However, learning Q-functions with Bellman error minimization typically
requires a Markovian state space.
When learning from partial observations such as images, the inputs might not be
Markovian.
\citet{dqn} introduced the concatenation of short observation sequences.
Prior methods \citep{drqn,oh2016minecraft,a3c,heess15rdpg}
have proposed to mitigate this issue by learning recurrent critics and Q-functions,
that depend on entire histories of observations.
The recurrent deterministic policy gradient \citep{heess15rdpg} for partially
observable control uses a similar DDPG-style estimator as used in ARM
(see Section \ref{sec:arm_impl}).
However, all of these changes increase the size of the input space,
increase variance, or make the optimization problem more complex.
Our method instead learns cumulative advantage functions that depend only on
the current observation, but can still handle non-Markovian problems.

The form of our advantage function update resembles positive temporal difference methods
\citep{peng16,van-hasselt07}.
Additionally, our update rule for a modified cumulative Q-function resembles
the average Q-function \citep{anschel17} used for variance reduction in Q-learning.
In both cases, the theoretical foundations of our method are based on cumulative regret minimization,
and the motivation is substantively different.
Previous work by \citet{ross11,ross+bagnell14} has connected
regret minimization to reinforcement learning, imitation learning,
and structured prediction,
although not with counterfactual regret minimization.
Regression regret matching \citep{waugh15} is based on a closely related idea,
which is to directly approximate the regret with a linear regression model,
however the use of a linear model is limited in representation compared to deep
function approximation.

Finally,
regret and value functions both typically take the form of expectations in
reinforcement learning.
An alternative view of value functions in RL is through the lens of
value distributions \citep{bellemare17dist}.
A connection between regret and distributional RL could lead to
very interesting future work.

\section{Advantage-based Regret Minimization}

In this section, we provide some background on
counterfactual regret minimization,
describe ARM in detail,
and give some intuition for why ARM works.

\subsection{Counterfactual Regret Minimization (CFR)}

First, we give a reinforcement learning-centric
exposition of counterfactual regret minimization (CFR)
\citep{zinkevich07,bowling15}.

The CFR model for partial observation is as follows.
Let $\mathcal S$ be the state space.
Let $\mathcal I$ be the space of \emph{information sets}:
an information set $I \in \mathcal I$ is a set of states $s \in I$,
where $s \in \mathcal S$,
such that only the information set $I$ is directly observable,
and the individual states contained in $I$ are hidden.
An information set $I$ is therefore a kind of \emph{aggregate state}.

In CFR, an extensive-form game is repeatedly played between $N$ players,
indexed by $i$.
We consider an iterative learning setting,
where at the $t$-th learning iteration,
the $i$-th player follows a fixed policy $\pi_t^i$,
choosing actions $a \in \mathcal A$
conditioned on information sets $I \in \mathcal I$
according to their policy $\pi_t^i(a|I)$.
Let $\sigma_t$ denotes the players' joint policy (their strategy profile).
For any strategy profile $\sigma$,
the $i$-th player has an expected value for playing the game,
$J^i(\sigma)$.
Let $\sigma_t^{-i}$ represent the joint policy of all players except the
$i$-th player;
similarly, let the tuple $(\pi_t^i,\sigma_t^{-i})$ be the product of
the $i$-th player's policy $\pi_t^i$ and
the other players' joint policy $\sigma_t^{-i}$.

The $i$-th player's \emph{overall regret} after $T$ learning iterations
is defined:
\begin{align}
  (R_T^i)^\text{(overall)}
  &= \max_{\pi_*^i} \sum_{t=1}^T J^i(\pi_*^i,\sigma_t^{-i}) - J^i(\pi_t^i,\sigma_t^{-i}).
  \label{eq:oregret}
\end{align}
What is the interpretation of the overall regret?
It is essentially how much better the $i$-th player could have done,
had it always followed an optimal policy in hindsight instead of the actual
sequence of policies $\pi_t^i$ it executed.
Intuitively, the difference $J^i(\pi_*^i,\sigma_t^{-i}) - J^i(\pi_t^i,\sigma_t)$
is the suboptimality of $\pi_t^i$ compared to $\pi_*^i$,
and the sum of the suboptimalities over learning iterations
yields the area inside the learning curve.
In other words, a smaller overall regret implies better sample efficiency.

Let $Q_{\sigma_t}^i(I,a)$ be the \emph{counterfactual value}
of the $i$-th player,
where the $i$-th player is assumed to reach $I$
and always chooses the action $a$ in the aggregate state $I$,
and otherwise follows the policy $\pi_t^i$,
while all other players follow the strategy profile $\sigma_t^{-i}$.
Similarly, let the expected counterfactual value be calculated as
$V_{\sigma_t}^i(I) = \sum_{a \in \mathcal A} \pi_t^i(a|I) Q_{\sigma_t}^i(I,a)$.
Let $(R_T^i)^\text{(CF)}(I,a)$ be the \emph{counterfactual regret}
of the $i$-th player,
which is the sum
of the advantage-like quantities $Q_{\sigma_t}^i(I,a) - V_{\sigma_t}^i(I)$
after $T$ learning iterations:
\begin{align}
  (R_T^i)^\text{(CF)}(I,a)
  &= \sum_{t=1}^T Q_{\sigma_t}^i(I,a) - V_{\sigma_t}^i(I) .
  \label{eq:cfregret_game}
\end{align}
Similarly, the \emph{immediate counterfactual regret} can be obtained from the
counterfactual regret by maximization over the player's action space:
\begin{align}
  (R_T^i)^\text{(immCF)}(I)
  &= \max_{a \in \mathcal A} \sum_{t=1}^T Q_{\sigma_t}^i(I,a) - V_{\sigma_t}^i(I) .
  \label{eq:icfregret}
\end{align}
The immediate counterfactual regret (Equation (\ref{eq:icfregret}))
possesses a similar interpretation
to the overall regret (Equation (\ref{eq:oregret})),
except that the immediate counterfactual regret and its constituent terms are
additionally functions of the aggregate state $I$.

Suppose that one were to briefly consider each aggregate state $I$ as a
separate, independent subproblem.
By naively treating the counterfactual regret $(R_T^i)^\text{(CF)}(I,a)$
for each $I$
as analogous to regret in an online learning setting,
then at each learning iteration one may simply plug the counterfactual regret
into a regret matching policy update \citep{rm}:
\begin{align}
  (\pi_{t+1}^i)^\text{RM}(a|I)
  &= \frac{\max( 0 , (R_t^i)^\text{(CF)}(I,a) )}{\sum_{a' \in \mathcal A} \max( 0 , (R_t^i)^\text{(CF)}(I,a') )}.
  \label{eq:rm_policy_game}
\end{align}
In the online learning setting,
the regret matching policy achieves regret that increases with
rate $O(\sqrt{T})$ in the number of iterations $T$
\citep{cesa03,gordon07};
we can then say that the regret matching policy is \emph{regret minimizing}.

It turns out that updating players' policies in the extensive game setting
by iteratively minimizing the immediate counterfactual regrets according to
Equation (\ref{eq:rm_policy_game})
will also minimize the overall regret $(R_T^i)^\text{(overall)}$
with upper bound $O(|\mathcal I| \sqrt{|\mathcal A| T})$.
The overall regret's $O(\sqrt{T})$ dependence on the number of iterations $T$
is not impacted by the structure of the information set space $\mathcal I$,
which is why CFR can be said to be robust to a certain kind of
partial observability.
This result forms the basis of the counterfactual regret minimization algorithm
and was proved by \citet{zinkevich07}.

Since we are interested in the application of CFR to
reinforcement learning,
we can write down ``1-player'' versions of the components above:
the counterfactual value,
reinterpreted as a \emph{stationary} state-action value function
$Q_{\pi|I \mapsto a}(I,a)$,
where the action $a$ is always chosen in the aggregate state $I$,
and the policy $\pi$ is otherwise followed
\citep{bellemare16};
the counterfactual regret, including its recursive definition:
\begin{align}
  R_T^\text{(CF)}(I,a)
  &= \sum_{t=1}^T Q_{\pi_t|I \mapsto a}(I,a) - V_{\pi_t|I}(I) \\
  &= R_{T-1}^\text{(CF)}(I,a) + Q_{\pi_T|I \mapsto a}(I,a) - V_{\pi_T|I}(I)
  \label{eq:cfregret}
\end{align}
where $V_{\pi_t|I}(I) = \sum_{a \in \mathcal A} \pi_t(a|I) Q_{\pi_t|I \mapsto a}(I,a)$;
and the regret matching policy update:
\begin{align}
  \pi_{t+1}^\text{RM}(a|I)
  &= \frac{\max( 0 , R_t^\text{(CF)}(I,a) )}{\sum_{a' \in \mathcal A} \max( 0 , R_t^\text{(CF)}(I,a') )}.
  \label{eq:rm_policy}
\end{align}

\subsection{CFR+}

CFR+ \citep{tammelin14} consists of a modification to CFR,
in which instead of calculating the full counterfactual regret as in
Equation (\ref{eq:cfregret}),
instead the counterfactual regret is recursively positively clipped:
\begin{align}
  \label{eq:clipped_counterfactual_regret}
  & R_T^\text{(CF+)}(I,a) \\
  &= [R_{T-1}^\text{(CF+)}(I,a)]_+ + Q_{\pi_T|I \mapsto a}(I,a) - V_{\pi_T|I}(I) \nonumber
\end{align}
where $[x]_+ = \max(0,x)$ is the positive clipping operator.
Comparing Equation (\ref{eq:cfregret}) with
Equation (\ref{eq:clipped_counterfactual_regret}),
the only difference in CFR+ is that the previous iteration's
quantity is positively clipped in the recursion.
This simple change turns out to yield a large practical improvement in
the performance of the algorithm \citep{bowling15}.
One intuition for why the positive clipping of CFR+ can improve upon CFR
is that because $[R_{T-1}^\text{(CF+)}]_+$ is nonnegative,
it provides a kind of ``optimism under uncertainty,''
adding a bonus to some transitions while ignoring others.
The CFR+ update has also been shown to do better than CFR when the best
action in hindsight changes frequently \citep{tammelin15}.
In the rest of this work we will solely build upon the CFR+ update.

\subsection{From CFR and CFR+ to ARM}
\label{sec:arm_impl}

Recall that CFR and CFR+ model partial observation in extensive games
with a space $\mathcal I$ of information sets or aggregate states.
Because we are interested in general observation spaces $\mathcal O$ in
partially observed Markov decision processes (MDPs),
we naively map CFR and CFR+ onto MDPs and replace $\mathcal I$ with
$\mathcal O$.
It is known that Markov or stochastic games can be converted to
extensive form \citep{littman94,sandholm12,kroer14},
and an MDP is a 1-player Markov game.
Because we are also interested in high dimensional observations such as images,
we estimate a counterfactual regret function approximation
\citep{waugh15}
parameterized as a neural network.

The exact form of the counterfactual regret
(either Equation (\ref{eq:cfregret})
or Equation (\ref{eq:clipped_counterfactual_regret})),
after translating aggregate states $I$ to observations $o$,
utilizes a stationary action-value function $Q_{\pi|o \mapsto a}(o,a)$.
It is unclear how to learn a truly stationary action-value function,
although \citet{bellemare16} propose a family of consistent Bellman operators
for learning locally stationary action-value functions.
We posit that the approximation
$Q_{\pi|o \mapsto a}(o,a) \approx Q_{\pi}(o,a)$,
where $Q_\pi(o,a)$ is the usual action-value function,
is acceptable when observations are rarely seen more than once in a typical
trajectory.

The above series of approximations finally yields a learnable function
using deep reinforcement learning;
this function is a ``clipped cumulative advantage function'':
\begin{align}
  A^+_T(o,a)
  &= \max(0,A^+_{T-1}(o,a)) + A_{\pi_T}(o,a)
  \label{eq:adv_recurrence}
\end{align}
where $A_\pi(o,a)$ is the usual advantage function.
Advantage-based regret minimization (ARM) is then the resulting batch-mode
deep RL algorithm that updates the policy to the regret matching distribution
on the cumulative clipped advantage function:
\begin{align}
  \pi_{t+1}(a|o)
  &= \frac{\max(0, A^+_{t}(o,a))}{\sum_{a' \in \mathcal A} \max(0, A^+_{t}(o,a'))}.
  \label{eq:adv_match}
\end{align}
At the $t$-th batch iteration of ARM,
a batch of data is collected by sampling trajectories using the
current policy $\pi_t$,
followed by two processing steps:
(a) fit $A^+_t$ using Equation (\ref{eq:adv_recurrence}),
then (b) set the next iteration's policy $\pi_{t+1}$ using
Equation (\ref{eq:adv_match}).

Below, we will use the subscript $k$ to refer to a timestep within a trajectory,
while the subscript $t$ refers to the batch iteration.
To implement Equation (\ref{eq:adv_recurrence}) with deep function
approximation,
we define two value function approximations, $V_{\pi_t}(o_k;\theta_t)$
and $Q^+_{t}(o_k,a_k;\omega_t)$,
as well as a target value function $V'(o_k;\varphi)$,
where $\theta_t$, $\omega_t$, and $\varphi$ are the learnable parameters.
The cumulative clipped advantage function is represented as
$A^+_t(o_k,a_k) = Q^+_t(o_k,a_k;\omega_t) - V_{\pi_t}(o_k;\theta_t)$.
Within each sampling iteration, the value functions are fitted using stochastic
gradient descent by sampling minibatches and performing gradient steps.
In practice, we use Adam to perform the optimization
\cite{adam}.
The state-value function $V_{\pi_t}(o_k;\theta_t)$
is fit using $n$-step returns with a moving target value function
$V'(o_{k+n};\varphi)$,
essentially using the estimator of the deep deterministic policy gradient
(DDPG) \citep{ddpg}.
In the same minibatch,
$Q^+_{t}(o_k,a_k;\theta_t)$ is fit to a similar loss, but with an
additional target reward bonus that incorporates the previous iteration's
cumulative clipped advantage, $\max( 0 , A^+_{t-1}(o_k,a_k) )$.
The regression targets $v_k$ and $q_k^+$ are defined in terms of
the $n$-step returns
$g_k^n = \sum_{k'=k}^{k+n-1} \gamma^{k'-k} r_{k'} + \gamma^n V'(o_{k+n};\varphi)$:
\begin{align}
  v_k
  &\triangleq g_k^n
  \label{eq:target_v}
  \\
  q_k
  &\triangleq r_k + \gamma g_{k+1}^{n-1}
  \\
  \phi_k
  &\triangleq Q^+_{t-1}(o_k,a_k;\omega_{t-1}) - V_{\pi_{t-1}}(o_k;\theta_{t-1})
  \\
  q_k^+
  &\triangleq \max( 0 , \phi_k ) + q_k.
  \label{eq:target_q}
\end{align}
Altogether, each minibatch step of the optimization subproblem consists of the
following three parameter updates:
\begin{align}
  \theta_t^{(\ell+1)}
  &\gets \theta_t^{(\ell)} - \frac{\alpha}{2} \nabla_{\theta_t^{(\ell)}} (V_{\pi_t}(o_k;\theta_t^{(\ell)}) - v_k)^2
  \label{eq:update_v} \\
  \omega_t^{(\ell+1)}
  &\gets \omega_t^{(\ell)} - \frac{\alpha}{2} \nabla_{\omega_t^{(\ell)}} (Q^+_{t}(o_k,a_k;\omega_t^{(\ell)}) - q_k^+)^2
  \label{eq:update_q} \\
  \varphi^{(\ell+1)}
  &\gets \varphi^{(\ell)} + \tau(\theta_t^{(\ell+1)} - \varphi^{(\ell)}).
  \label{eq:update_target}
\end{align}
The advantage-based regret minimization algorithm
is summarized in Algorithm \ref{alg:arm}.

We note again that we use a biased value function estimator,
whereas only the full returns are guaranteed to be unbiased in
non-Markovian settings.
In the high-dimensional domains we evaluated on,
the most practical choice of deep advantage or value function approximation is
based on biased but lower variance estimation,
such as the $n$-step returns we use in practice
\citep{trpo_gae,gu17_ipg}.
We also note that ARM is essentially CFR with different design choices
to facilitate function approximation.
CFR is proven to converge in domains with non-Markovian information set spaces
when using tabular representations,
suggesting that our use of biased advantage estimation does not fundamentally
preclude the applicability or effectiveness of ARM on non-Markovian domains.

\begin{algorithm}
  \caption{Advantage-based regret minimization (ARM).}
  \begin{algorithmic}
    \STATE{initialize $\pi_1 \gets \text{uniform}$, $\theta_{0} , \omega_{0} \gets \text{arbitrary}$}
    \FOR{$t$ in $1, \ldots$}
      \STATE{collect batch of trajectory data $\mathcal D_t \sim \pi_t$}
      \STATE{initialize $\theta_t \gets \theta_{t-1}$, $\omega_t \gets \omega_{t-1}$, $\varphi \gets \theta_{t-1}$}
      \FOR{$\ell$ in $0, \ldots$}
        \STATE{sample transitions:}
        \STATE{$(o_k,a_k,r_k,o_{k+1}) \sim \mathcal D_t$}
        \STATE{$\delta_{k} \gets 1 - \mathbb I[o_{k}~\text{is terminal}]$}
        \STATE{calculate $n$-step returns:}
        \STATE{$g_k^n \gets \sum_{k'=k}^{k+n-1} \gamma^{k'-k} r_{k'} + \gamma^n \delta_{k+n} V'(o_{k+n};\varphi)$}
        \STATE{calculate target values:}
        \IF{$t = 1$}
          \STATE{$\phi_k \gets 0$}
        \ELSE
          \STATE{$\phi_k \gets Q^+_{{t-1}}(o_k,a_k;\omega_{t-1}) - V_{\pi_{t-1}}(o_k;\theta_{t-1})$}
        \ENDIF
        \STATE{$v(o_k) \gets g_k^n$}
        \STATE{$q^+(o_k,a_k) \gets \max(0,\phi_k) + r_k + \gamma g_{k+1}^{n-1}$}
        \STATE{update $\theta_t$ (Equation (\ref{eq:update_v}))}
        \STATE{update $\omega_t$ (Equation (\ref{eq:update_q}))}
        \STATE{update $\varphi$ (Equation (\ref{eq:update_target}))}
      \ENDFOR
      \STATE{set $\pi_{t+1}(a|o) \propto \max(0, Q^+_{t}(o,a;\omega_t) - V_{\pi_t}(o;\theta_t))$}
    \ENDFOR
  \end{algorithmic}
  \label{alg:arm}
\end{algorithm}

\subsection{ARM vs.~Existing Policy Gradient Methods}
\label{sec:arm_vs_pg}

We note that the fundamental operation in CFR-like algorithms, including ARM,
is exemplified by the counterfactual regret update in
Equation (\ref{eq:cfregret})
which superficially looks quite similar to a policy gradient-style update:
the update is in the direction of a Q-value minus a baseline.
However, we can take the analogy between ARM and policy gradient methods
further and show that
ARM represents an inherently different update compared to existing
policy gradient methods.

Recent work has shown that policy gradient methods and Q-learning methods are
connected via maximum entropy RL
\citep{odonoghue17,haarnoja17,nachum17,schulman17}.
One such perspective is from
the soft policy iteration framework for batch-mode reinforcement learning
\citep{haarnoja18},
where at the $t$-th batch iteration
the updated policy is obtained by minimizing the average KL-divergence
between the parametric policy class $\Pi$ and a target policy $f_t$.
Below is the soft policy iteration update,
where the subscript $t$ refers to the batch iteration:
\begin{align}
  & \pi_{t+1} \nonumber
  = \arg \min_{\pi \in \Pi} \mathbb E_{o \sim \rho_t} [ D_{\text{KL}}( \pi(\cdot|o) \| f_t(\cdot|o) ) ] \\
  &= \arg \min_{\pi \in \Pi} \mathbb E_{o \sim \rho_t , a \sim \pi(\cdot|o)} [ \log( \pi(a|o) ) - \log( f_t(a|o) ) ].
\end{align}
Using the connection between policy gradient methods and Q-learning,
we define the policy gradient target policy as
a Boltzmann or softmax distribution
on the entropy regularized advantage function $A^{\text{$\beta$-soft}}$:
\begin{align}
  f_t^{\text{PG}}(a|o)
  &\triangleq \frac{\exp( \beta A_t^{\text{$\beta$-soft}}(o,a) )}{\sum_{a' \in \mathcal A} \exp( \beta A_t^{\text{$\beta$-soft}}(o,a') )}.
\end{align}
Now, parameterizing the policy $\pi$ in terms of an explicit parameter $\theta$,
we obtain the expression for the existing policy gradient update,
where $b(o)$ is a baseline function:
\begin{align}
  \label{eq:pg_grad}
  \Delta \theta^{\text{PG}}
  &\propto \mathbb E_{o \sim \rho_t , a \sim \pi(\cdot|o;\theta)} \Big[ \nabla_\theta \log( \pi(o|a;\theta) ) \cdot \\
  &~~~~~~~~~~ \big( -(1/\beta) \log( \pi(o|a;\theta) ) \nonumber \\
  &~~~~~~~~~~~~ + A_t^{\text{$\beta$-soft}}(o,a) - b(o) \big) \Big]. \nonumber
\end{align}
The non-entropy-regularized policy gradient arises in the limit
$\beta \to \infty$, at which point $(1/\beta)$ vanishes.

Note that an alternative choice of target policy $f_t$ will lead to a different
kind of policy gradient update.
A policy gradient algorithm based on ARM instead proposes
a target policy based on the regret matching distribution:
\begin{align}
  f_t^{\text{ARM}}(a|o)
  &\triangleq \frac{\max( 0 , A_t^+(o,a) )}{\sum_{a' \in \mathcal A} \max( 0 , A_t^+(o,a') )}.
\end{align}
Similarly, we can express the ARM-like policy gradient update, where again
$b(o)$ is a baseline:
\begin{align}
  \label{eq:arm_pg_grad}
  \Delta \theta^{\text{ARM}}
  &= \mathbb E_{o \sim \rho_t , a \sim \pi(\cdot|o;\theta)} \Big[ \nabla_\theta \log( \pi(o|a;\theta) ) \cdot \\
  &~~~~~~~~~~ \big( -\log( \pi(o|a;\theta) ) \nonumber \\
  &~~~~~~~~~~~~ + \log( \max( 0 , A_t^+(o,a) ) ) - b(o) \big) \Big]. \nonumber
\end{align}
Comparing Equations (\ref{eq:pg_grad}) and (\ref{eq:arm_pg_grad}),
we see that the ARM-like policy gradient (Equation (\ref{eq:arm_pg_grad}))
has a logarithmic dependence on the clipped advantage-like function
$\max(0,A^+)$,
whereas the existing policy gradient (Equation (\ref{eq:pg_grad})) is only
linearly dependent on the advantage function $A^{\text{$\beta$-soft}}$.
This difference in
logarithmic vs.~linear dependence is responsible for a large part of the
inherent distinction of ARM from existing policy gradient methods.
In particular, the logarithmic dependence in an ARM-like update
may be less sensitive to advantage overestimation
compared to existing policy gradient methods,
perhaps serving a similar purpose to the double Q-learning estimator
\citep{ddqn}
or consistent Bellman operators \citep{bellemare16}.

We also see that for the existing policy gradient (Equation (\ref{eq:pg_grad})),
the $-(1/\beta)\log( \pi(a|o;\theta) )$ term,
which arises from the policy entropy,
is vanishing for large $\beta$.
On the other hand, for the ARM-like policy gradient
(Equation (\ref{eq:arm_pg_grad})),
there is no similar vanishing effect on the equivalent policy entropy term,
suggesting that ARM may perform a kind of entropy regularization by default.

In practice we cannot implement an ARM-like policy gradient exactly as in
Equation (\ref{eq:arm_pg_grad}),
because the positive clipping $\max( 0 , A^+ )$
can yield $\log(0)$.
However we believe this is not an intrinsic obstacle and leave the
question of how to implement an ARM-like policy gradient to future work.

\subsection{Comparing ARM with Other Methods in Partially Observable Domains}
\label{sec:intuition}

The regret matching policy which is fundamental to ARM can be interpreted
as a more nuanced form of exploration compared to the epsilon-greedy policy
used with Q-learning.
In partially observable domains, the optimal policy may generally be
stochastic \cite{singh94}.
So $\epsilon$-greedy policies which put a substantial probability mass on
$\arg \max_a Q(o,a)$, e.g.~by setting $\epsilon=0.01$, can be suboptimal,
especially compared to more general distributions on discrete actions,
such as ARM's regret matching policy.
The softmax policy typically learned by policy gradient methods is also quite
general,
but can still put too much probability mass on one action without compensation
by an explicit entropy bonus
as done in maximum entropy reinforcement learning.

Policy gradient methods have an overall regret bound of
$R_T \le B^2/\eta + \eta G^2 T$
derived from stochastic gradient descent,
where $B$ is an upper bound on the policy parameter $\ell^2$-norm,
$G^2$ is an upper bound on the second moments of the stochastic
gradients,
and $\eta$ is the learning rate
\cite{dick15},
Assuming an optimal learning rate $\eta = B/(G\sqrt{T})$,
the policy gradient regret bound becomes
$R_T \le 2 B G \sqrt{T}$.
CFR has an overall regret bound of
$R_T \le \Delta |\mathcal I| \sqrt{|\mathcal A| T}$,
where $\Delta$ is the positive range of returns,
and $|\mathcal I|$ and $|\mathcal A|$ are cardinalities of the
information set space and action space, respectively
\cite{zinkevich07}.
Both regret bounds have $O(\sqrt{T})$ dependence on $T$.
As mentioned earlier, policy gradient methods can also be robust to
partial observation,
but unlike in the case of CFR,
the dependence of the policy gradient regret bound on an
optimal learning rate and on gradient variance may affect policy gradient
convergence in the presence of estimation error.
On the other hand, the constants in the CFR regret bound are constant
properties of the environment.

For Q-learning per se, we are not aware of any known regret bound.
Szepesv{\'a}ri proved that an upper bound on the convergence rate of Q-learning
(specifically, the convergence rate of $\|Q_T(s,a)-Q^*(s,a)\|_\infty$),
assuming a fixed exploration strategy,
depends on an exploration condition number,
which is the ratio of minimum to maximum state-action occupation frequencies
\citep{szepesvari98},
and which describes how ``balanced'' the exploration strategy is.
If partial observability leads to imbalanced exploration due to confounding
of states from perceptual aliasing \citep{mccallum97},
then Q-learning should be negatively affected in convergence and possibly
in absolute performance.

\section{Experiments}
\label{sec:experiments}

Because we hypothesize that ARM should perform well in partially observable
domains,
we conduct our experiments on visual tasks that naturally provide partial
observations of state.
Our evaluations use feedforward convnets with frame history inputs;
our hyperparameters are listed in Section A1 of the Supplementary Material.
We are interested in comparing ARM with other advantage-structured methods,
primarily:
(a) double deep Q-learning with dueling network streams \citep{ddqn,dueling},
which possesses an advantage-like parameterization of its Q-function
and assumes Markovian observations;
and (b) TRPO \citep{trpo,trpo_gae},
which estimates an empirical advantage using a baseline state-value function
and can handle non-Markovian observations.

\subsection{Learning First-person 3D Navigation}
\label{sec:doom}
\label{sec:minecraft}

We first evaluate ARM on the task of learning first-person navigation in
3D maze-like tasks from two domains:
ViZDoom \citep{vizdoom} based on the game Doom,
and Malm{\"o} \citep{malmo} based on the game Minecraft.
Doom and Minecraft both feature an egocentric viewpoint, 3D perspective, and
rich visual textures.
We expect that both domains exhibit a substantial degree of partial
observability since only the immediate field-of-view of the environment is
observable by the agent due to first-person perspective.

\begin{figure}[ht]
  \begin{center}
  \begin{tabular}{c@{\hspace{4pt}}c@{\hspace{4pt}}c}
    \includegraphics[width=0.28\linewidth]{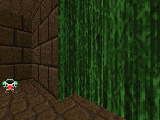}
  & \includegraphics[width=0.28\linewidth]{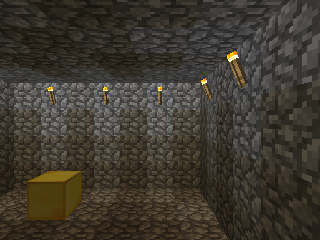}
  & \includegraphics[width=0.28\linewidth]{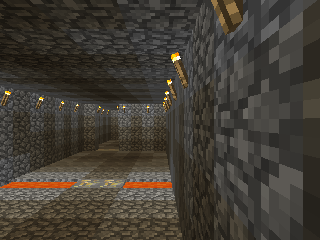}
  \end{tabular}
  \caption{
    Screenshots from
    (left) Doom MyWayHome,
    (middle) Minecraft level 1,
    and (right) Minecraft level 4.
  }
  \label{fig:show_doom}
  \end{center}
  \vskip -0.2in
\end{figure}

In Doom MyWayHome, the agent is randomly placed in one of several rooms
connected in a maze-like arrangement,
and the agent must reach an item that has a fixed visual appearance and is in
a fixed location before time runs out.
For Minecraft, we adopt the teacher-student curriculum learning task of
\citet{matiisen17},
consisting of 5 consecutive ``levels'' that successively increase the
difficulty of completing the simple task of reaching a gold block:
the first level (``L1'') consists of a single room;
the intermediate levels (``L2''--``L4'') consist of a corridor with lava-bridge
and wall-gap obstacles;
and the final level (``L5'') consists of a $2\times2$ arrangement of rooms
randomly separated by lava-bridge or wall-gap obstacles.
Examples of the MyWayHome and the Minecraft levels are shown in
Figure \ref{fig:show_doom}.

\begin{figure}[ht]
  \begin{center}
  \includegraphics[width=0.54\linewidth]{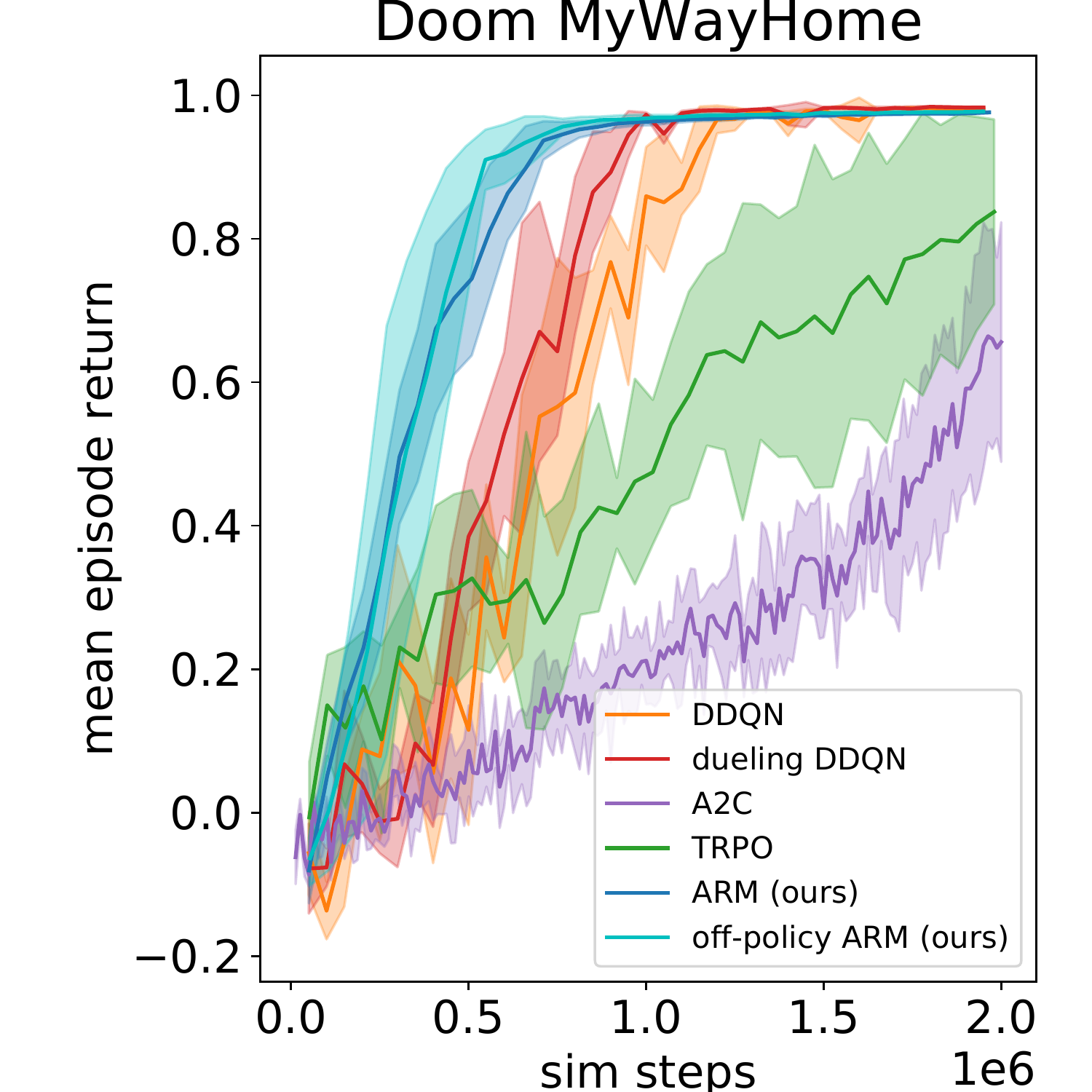}
  \caption{
    Evaluation on the Doom MyWayHome task.
  }
  \label{fig:doom}
  \end{center}
  \vskip -0.2in
\end{figure}

\begin{figure}[ht]
  \begin{center}
  \includegraphics[width=0.45\linewidth]{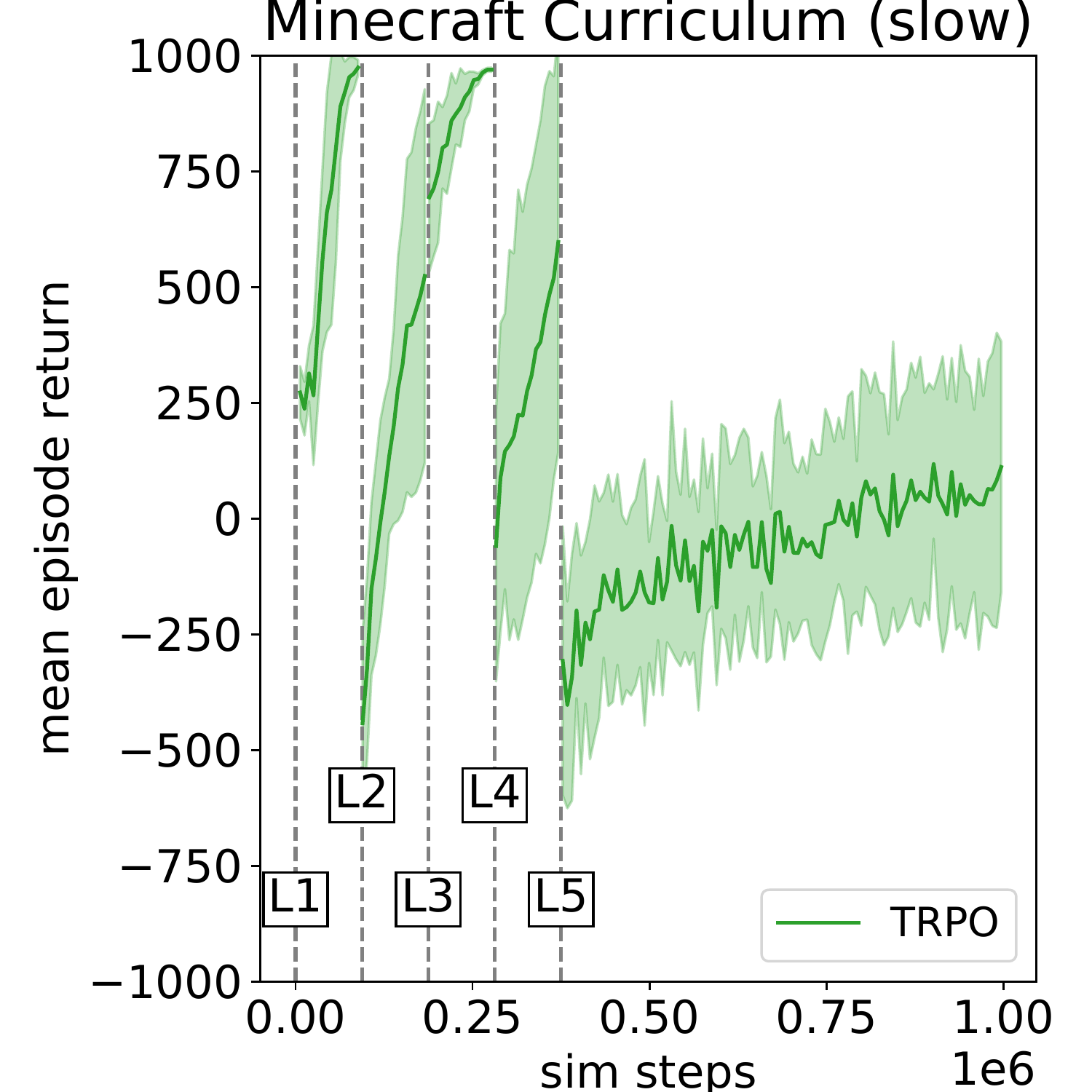}
  \includegraphics[width=0.45\linewidth]{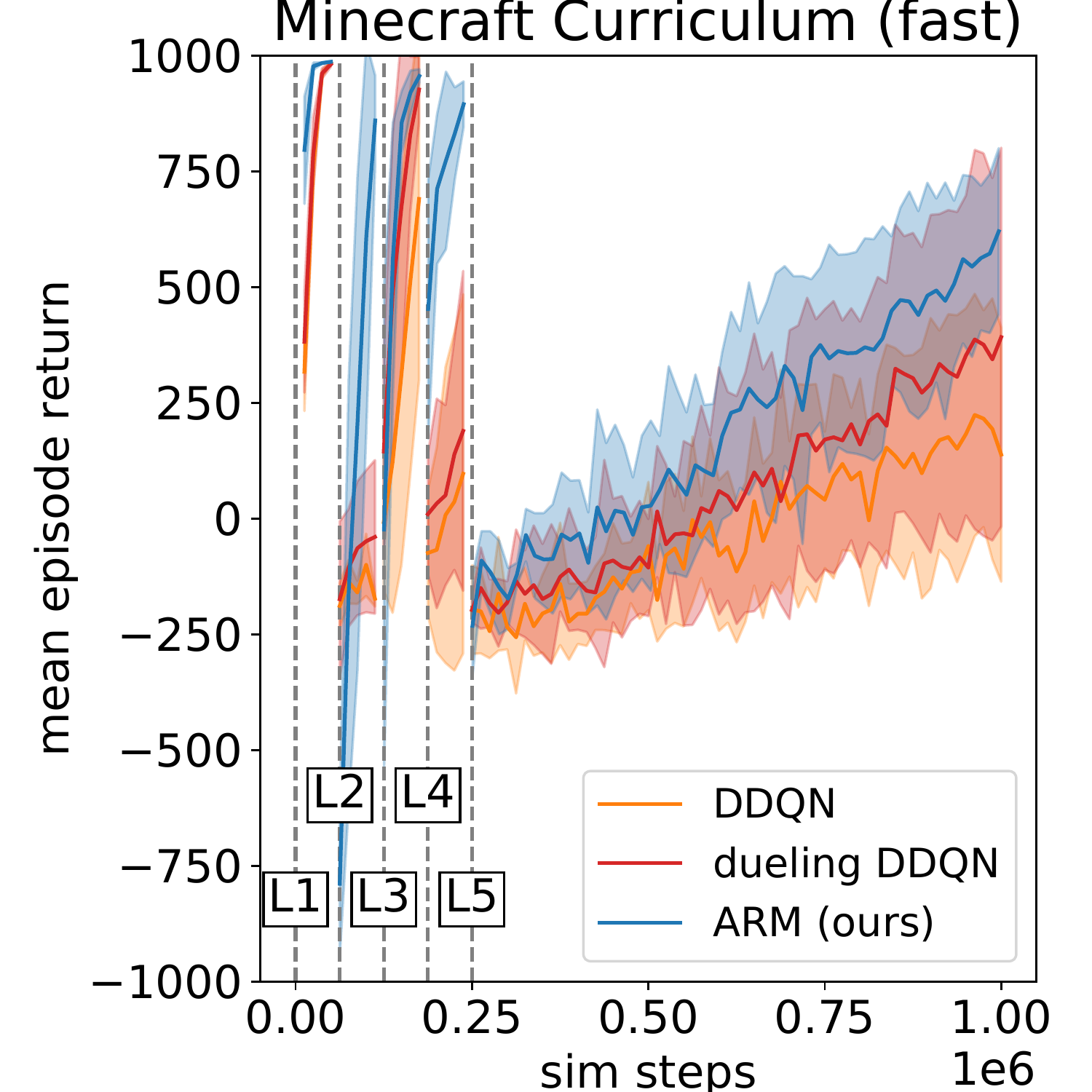}
  \caption{
    Evaluation on a Minecraft curriculum learning task.
    The simulator step counts at which each level begins are labeled and
    demarcated with dashed vertical lines.
  }
  \label{fig:minecraft}
  \end{center}
  \vskip -0.2in
\end{figure}

Our results on Doom and Minecraft are in Figures \ref{fig:doom} and \ref{fig:minecraft}.
Unlike previous evaluations which augmented raw pixel observations with extra
information about the game state, e.g.~elapsed time ticks or remaining health
\citep{vizdoom,dosovitskiy17},
in our evaluation we forced all networks to learn using only visual input.
Despite this restriction, ARM is still able to quickly learn policies with
minimal tuning of hyperparameters
and to reach close to the maximum achievable score in under 1 million simulator
steps,
which is quite sample efficient.
On MyWayHome, we observed that ARM generally learned a well-performing
policy more quickly than other methods.
Additionally, we found that ARM is able to take advantage of an off-policy
replay memory when learning on MyWayHome by storing the trajectories of previous
sampling batches and applying an importance sampling correction to the
$n$-step return estimator;
please see Section A2 in the Supplementary Material.

We performed our Minecraft experiments using fixed
curriculum learning schedules
to evaluate the sample efficiency of different algorithms:
the agent is initially placed in the first level (``L1''),
and the agent is advanced to the next level
whenever a preselected number of simulator steps have elapsed,
until the agent reaches the last level (``L5'').
We found that
ARM and dueling double DQN both were able to learn on an aggressive
``fast'' schedule of only 62500 simulator steps between levels.
TRPO required a ``slow'' schedule of 93750 simulator steps between levels
to reliably learn.
ARM was able to consistently learn a well performing policy on
all of the levels,
whereas
double DQN learned more slowly on some of the intermediate levels.
ARM also more consistently reached a high score on the final, most difficult
level (``L5'').

\subsection{Learning with Partially Observed Objects}
\label{sec:atari}

Previously in Section \ref{sec:intuition},
we argued that the convergence results for CFR suggest that reducing the
size of the observation space could improve the convergence of CFR compared to
methods based on policy gradients or Q-learning.
We would expect that by controlling the degree of partial observability
in a fixed task,
one could expect the relative sample efficiency or performance of ARM to
improve compared to other methods.

Whereas the navigation tasks in Section \ref{sec:doom} only dealt with
first-person motion in static 3D environments,
tasks that add other objects which may move and are not always visible to the
agent intuitively ought to be less observable.
To test the effect of the degree of partial observability on fixed tasks with
objects,
we conduct experiments with at least two variants of each task:
one experiment is based on the unmodified task,
but the other experiments
occlude or mask essential pixels in observed frames to raise the
difficulty of identifying relevant objects.

We evaluate ARM and other methods on two tasks with objects:
Doom Corridor via ViZDoom,
and Atari Pong via the Arcade Learning Environment \citep{ale}.
In Corridor, the agent is placed in a narrow 3D hallway and must avoid getting
killed by any of several shooting monsters scattered along the route
while trying to reach the end of the hallway.
In Pong, the agent controls a paddle and must hit a moving ball past the
computer opponent's paddle on the other side of the screen.

\begin{figure}[ht]
  \begin{center}
  \begin{tabular}{c@{\hspace{4pt}}c@{\hspace{4pt}}c@{\hspace{4pt}}c}
    \includegraphics[trim={0 15px 0 34px},clip,width=0.18\linewidth]{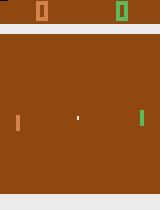}
  & \includegraphics[trim={0 15px 0 34px},clip,width=0.18\linewidth]{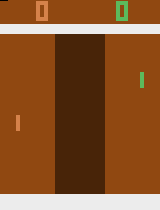}
  & \includegraphics[width=0.24\linewidth]{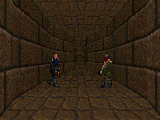}
  & \includegraphics[width=0.24\linewidth]{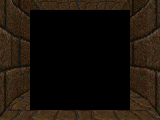}
  \end{tabular}
  \caption{
    Screenshots from Atari Pong (left two)
    and Doom Corridor+ (right two),
    unmodified and with occlusion.
  }
  \label{fig:show_pong}
  \end{center}
  \vskip -0.1in
\end{figure}

In our implementation of Corridor, which we call ``Corridor+,''
we restrict the action space of the agent to only be able
to turn left or right and to move to one side.
Because the agent spawns facing the end of the hallway,
the agent has to learn to first turn to its side to orient itself
sideways before moving,
while also trying to avoid getting hit from behind by monsters in
close proximity.
To induce partial observability in Corridor,
we mask out a large square region in the center of the frame buffer.
An example of this occlusion is shown in Figure \ref{fig:show_pong}.
Because the only way for the agent to progress toward the end of the hallway
is by moving sideways,
the monsters that are closest to the agent,
and hence the most dangerous,
will appear near the center of the agent's field-of-view.
So the center pixels are the most natural ones to occlude in Corridor+.

For Pong, we choose a rectangular area in the middle of the frame between the
two players' paddles,
then we set all pixels in the rectangular area to the background color.
An illustration of the occlusion is shown in Figure \ref{fig:show_pong}.
When the ball enters the occluded region, it completely disappears from view so
its state cannot be reconstructed using a limited frame history.
Intuitively, the agent needs to learn to anticipate the trajectory of the ball
in the absence of visual cues of the ball's position and velocity.

It is also known that for Atari games in general,
with only 4 observed frames as input,
it is possible to predict hundreds of frames into the future on some games
using only a feedforward dynamics model
\citep{oh2015action}.
By default, all of our networks receive as input a frame history of length 4.
This suggests that limiting the frame history length in Pong is another
effective perturbation for reducing observability.
When the frame history length is limited to one,
then reconstructing the velocity of moving objects in Pong becomes
more difficult \citep{drqn}.

\begin{figure}[ht]
  \begin{center}
  \includegraphics[width=0.45\linewidth]{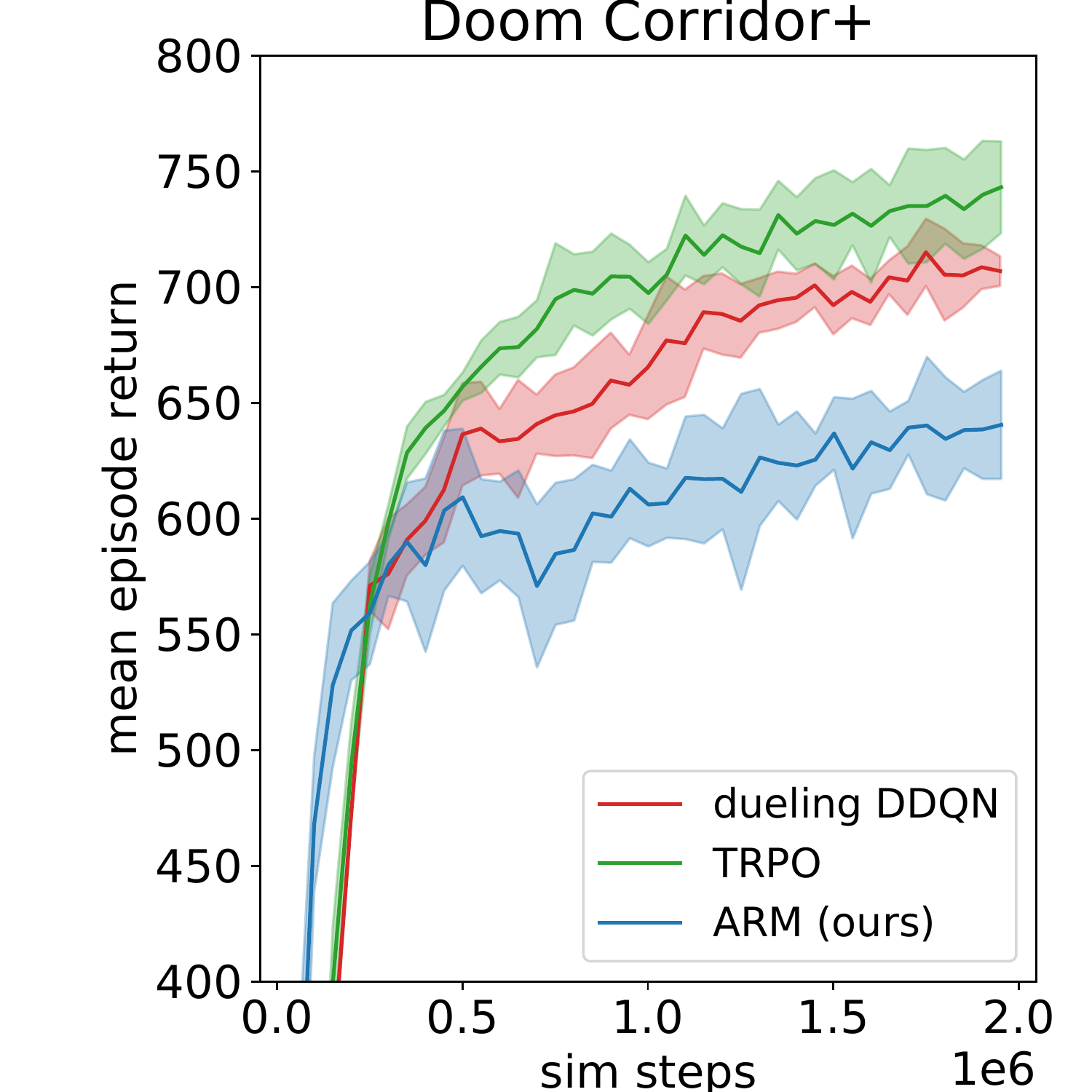}
  \includegraphics[width=0.45\linewidth]{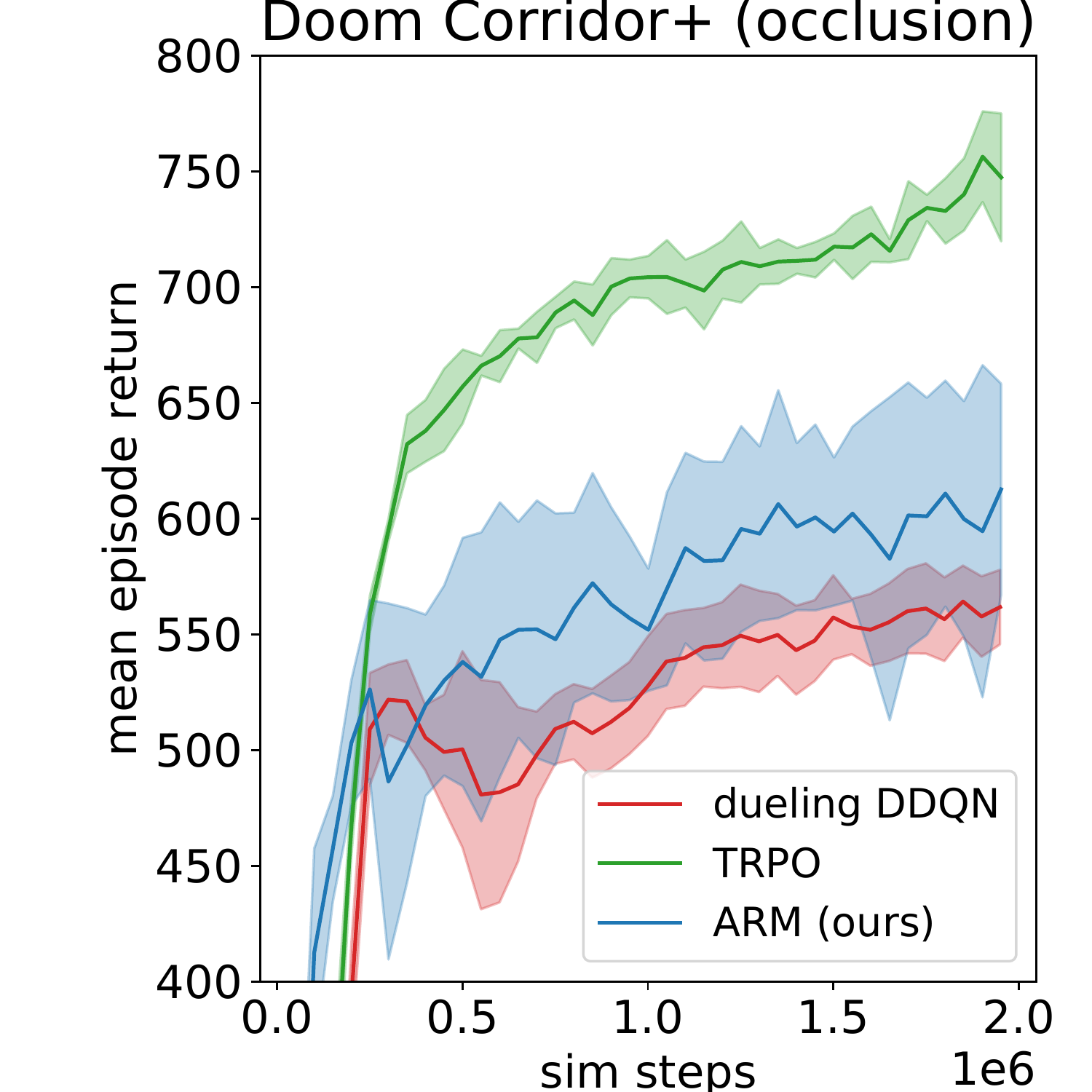}
  \caption{
    Results on Doom Corridor+:
    (left) unmodified,
    (right) with occlusion.
  }
  \label{fig:corridor_results}
  \end{center}
  \vskip -0.1in
\end{figure}

Our results on Corridor+ are shown in Figure \ref{fig:corridor_results}.
Among all our experiments, it is on Corridor+ that TRPO performs the
best compared to all other methods.
One distinguishing feature of TRPO is its empirical full return estimator,
which is unbiased on non-Markovian observations,
but which can have greater variance than the $n$-step return estimator used in
deep Q-learning and ARM.
On Corridor+, there appears to be a benefit to using the
unbiased full returns over the biased $n$-step returns,
which speaks to the non-Markovian character of the Corridor+ task.
It is also evident on Corridor+ that the performance of
deep Q-learning suffers when the observability is reduced
by occlusion,
and that by comparison ARM is relatively unaffected,
despite both methods using biased $n$-step returns with the same value of $n$
($n=5$).
This suggests that even when ARM is handicapped by the bias of its return
estimator,
ARM is intrinsically more robust to the non-Markovian observations that arise
from partial observability.
One possible direction to improve ARM is through
alternative return estimators \citep{trpo_gae,retrace}.

\begin{figure}[!ht]
  \begin{center}
  \includegraphics[width=0.32\linewidth]{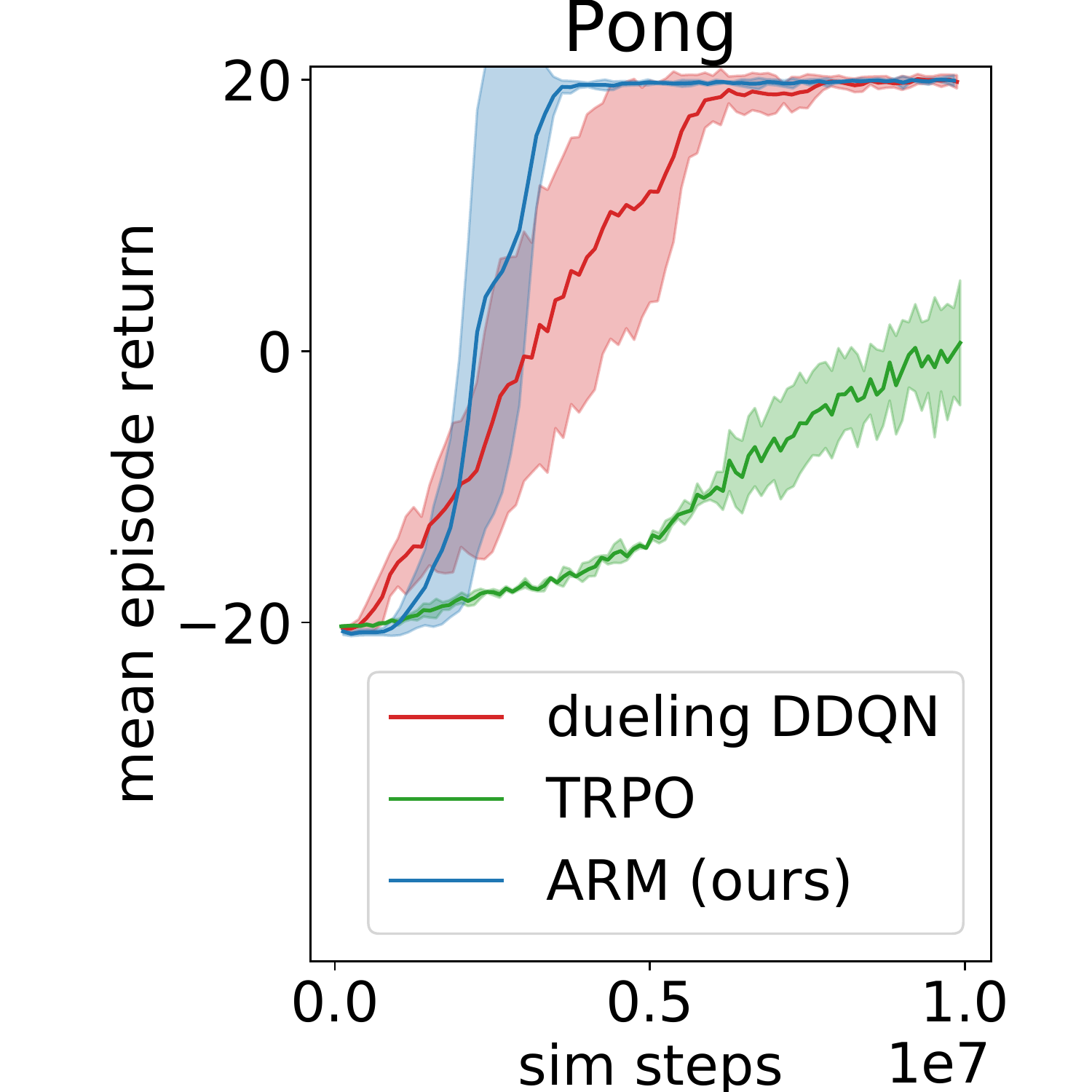}
  \includegraphics[width=0.32\linewidth]{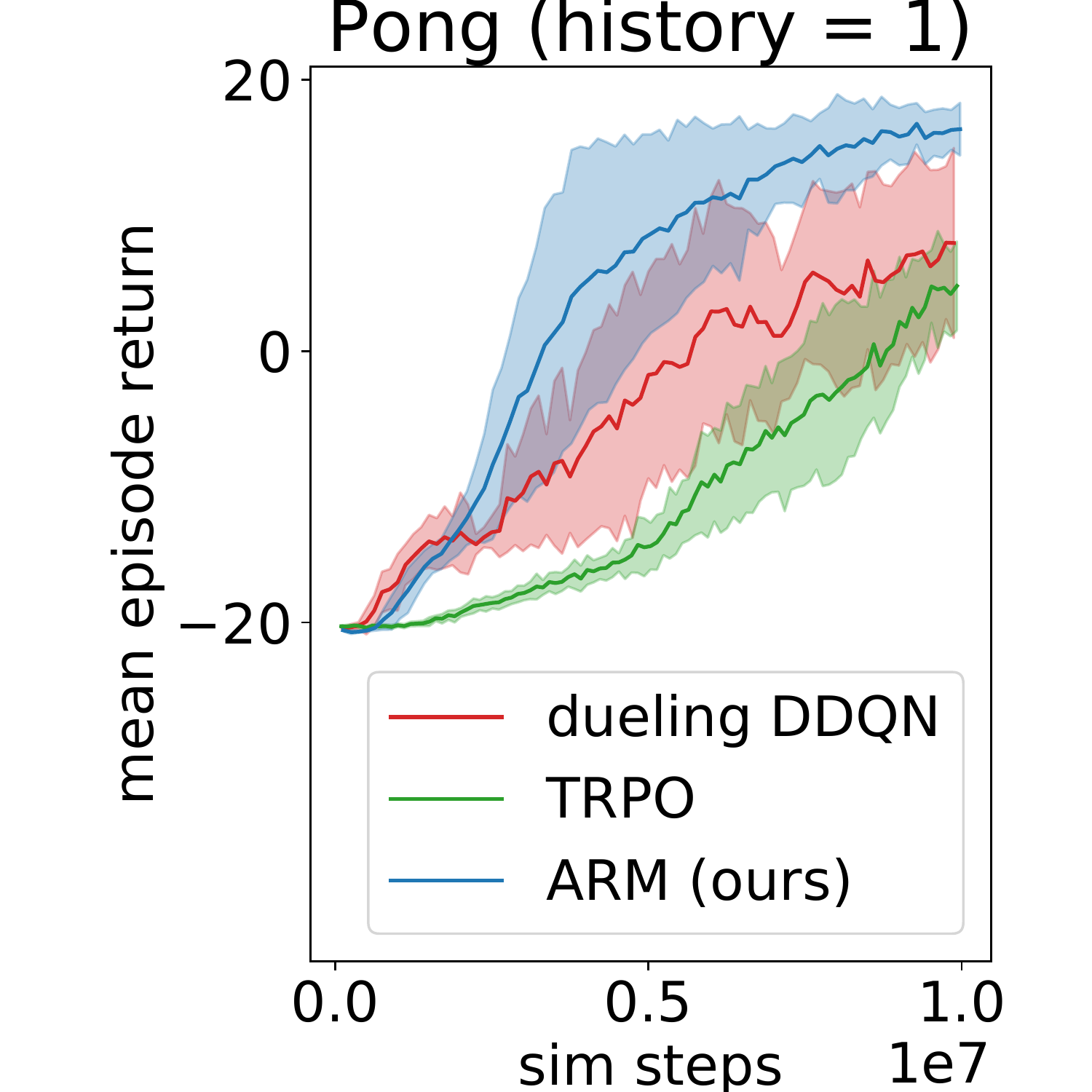}
  \includegraphics[width=0.32\linewidth]{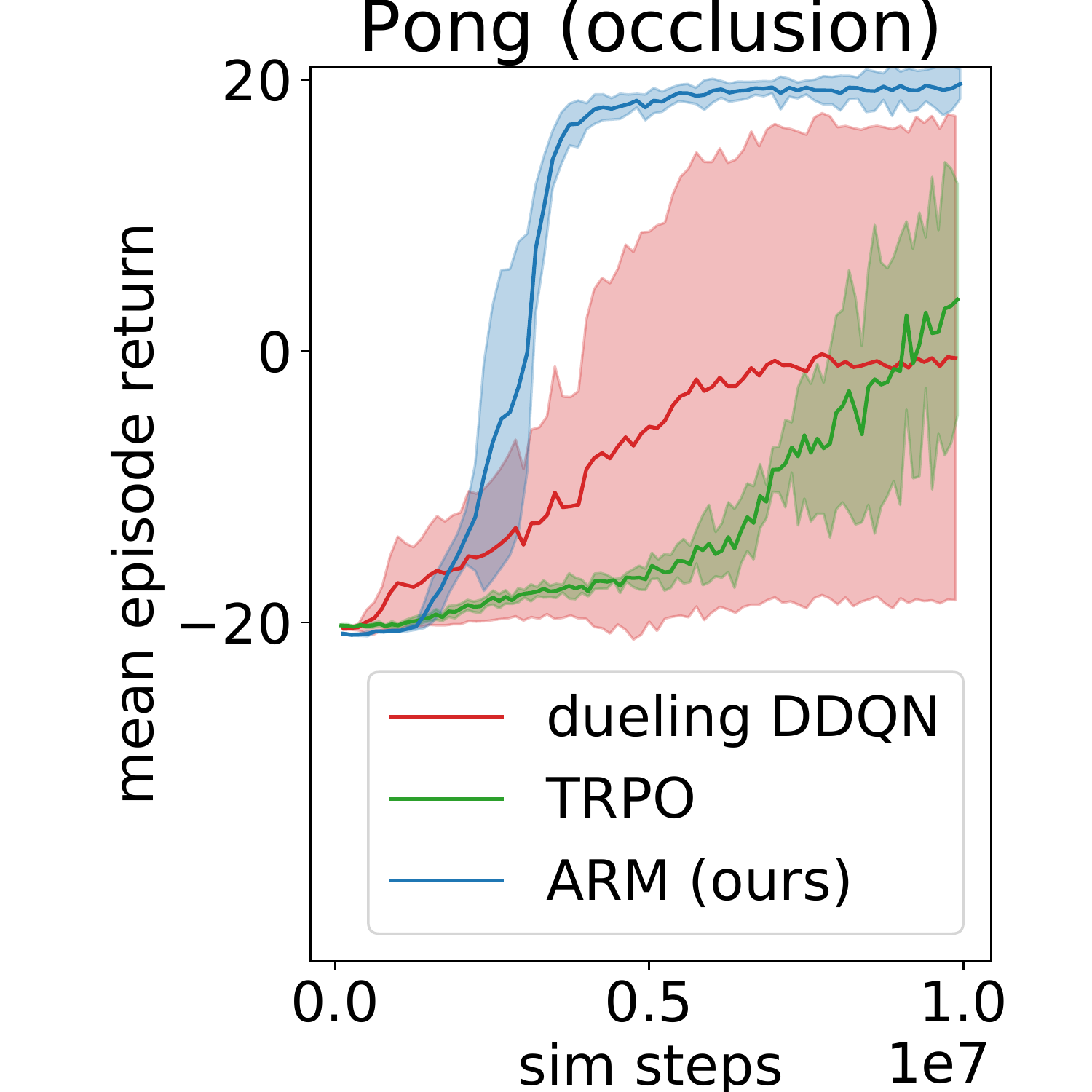}
  \caption{
    Results on Atari Pong:
    (left) unmodified,
    (middle) with limited frame history,
    (right) with occlusion.
  }
  \label{fig:pong_results}
  \end{center}
  \vskip -0.1in
\end{figure}

Our results on Pong are shown in Figure \ref{fig:pong_results}.
The convergence of ARM on all three variants of Pong suggests that ARM is not
affected much by partial observation in this domain.
As expected, when observability is reduced in
the limited frame history and occlusion experiments,
deep Q-learning performs worse and converges more slowly than ARM.
TRPO is generally less sample efficient than either ARM or DQN,
although like ARM it seems to be resilient to the partial observation
perturbations.

\section{Discussion}

In this paper, we presented a novel deep reinforcement learning algorithm
based on counterfactual regret minimization (CFR).
We call our method advantage-based regret minimization (ARM).
Similarly to prior methods that learn state or state-action value functions,
our method learns a cumulative clipped advantage function of observation and
action.
However, in contrast to these prior methods, ARM is well suited to partially
observed or non-Markovian environments,
making it an appealing choice in a number of difficult domains.
When compared to baseline methods, including deep Q-learning and TRPO,
on partially observable tasks
such as first-person navigation in Doom and Minecraft
and interacting with partially observed objects in Doom and Pong,
ARM is robust to the degree of partial observability and
can achieve substantially better sample efficiency and performance.
This illustrates the value of ARM for partially observable problems.
In future work, we plan to further explore applications of ARM to more complex
tasks, including continuous action spaces.

\section*{Acknowledgements}

We thank Tambet Matiisen for providing the Malm{\"o} curriculum learning tasks,
Deepak Pathak for help with ViZDoom experiments,
and Tuomas Haarnoja, Richard Li, and Haoran Tang for helpful discussions.
This work was partially supported by
Berkeley DeepDrive,
ADEPT Lab industrial sponsors and affiliates Intel, Google, Siemens, and SK Hynix,
DARPA Award Number HR0011-12-2-0016,
and ASPIRE Lab industrial sponsors and affiliates Intel, Google, HPE, Huawei,
LGE, Nokia, NVIDIA, Oracle, and Samsung.

\bibliography{arm}
\bibliographystyle{icml2018new}

\newpage

\setcounter{section}{0}
\renewcommand{\thesection}{A\arabic{section}}

\section{Experimental Details}

For each experiment, we performed 5 trials for each method
(some TRPO experiments on Pong are for 3 trials).

\subsection{Hyperparameters}

Our hyperparameter choices for each method are listed below.
For all methods using Adam, we roughly tuned the learning rate to find the
largest that consistently converged in unmodified variants of tasks.

\subsubsection{ARM}

Please see Tables \ref{tab:arm} and \ref{tab:offarm} for hyperparameters
used with ARM.
Note that our choice of ARM hyperparameters yields an equivalent number
of minibatch gradient steps per sample as used by deep Q-learning,
i.e.~1 Adam minibatch gradient step per 4 simulator steps;
c.f.~Table \ref{tab:dqn} for the deep Q-learning hyperparameters.
We kept hyperparameters
(other than the learning rate and the number of steps $n$)
constant across tasks.

\begin{table}[h]
\caption{
  Hyperparameters for ARM.
}
\label{tab:arm}
\vskip 0.15in
\begin{center}
\begin{small}
\begin{sc}
\begin{tabular}{lrrr}
\toprule
Hyperparameter          & Atari                 & Doom                  & Minecraft \\
\midrule
Adam learning rate      & $1\mathrm{e}^{-4}$    & $1\mathrm{e}^{-5}$    & $1\mathrm{e}^{-5}$ \\
Adam minibatch size     & $32$                  & $32$                  & $32$ \\
batch size              & $12500$               & $12500$               & $12500$ \\
gradient steps          & $3000$                & $3000$                & $3000$ \\
moving average ($\tau$) & $0.01$                & $0.01$                & $0.01$ \\
$n$-steps               & $1$                   & $5$                   & $5$ \\
\bottomrule
\end{tabular}
\end{sc}
\end{small}
\end{center}
\vskip -0.1in
\end{table}

\begin{table}[h]
\caption{
  Hyperparameters for off-policy ARM.
}
\label{tab:offarm}
\vskip 0.15in
\begin{center}
\begin{small}
\begin{sc}
\begin{tabular}{lr}
\toprule
Hyperparameter          & Doom \\
\midrule
Adam learning rate      & $1\mathrm{e}^{-5}$ \\
Adam minibatch size     & $32$ \\
batch size              & $1563$ \\
gradient steps          & $400$ \\
importance weight clip  & $1$ \\
moving average ($\tau$) & $0.01$ \\
$n$-steps               & $5$ \\
replay memory max       & $25000$ \\
\bottomrule
\end{tabular}
\end{sc}
\end{small}
\end{center}
\vskip -0.1in
\end{table}

\subsubsection{A2C}

Please see Table \ref{tab:a2c} for hyperparameters used with A2C.
We found that increasing the number of steps $n$ used to calculate the
$n$-step returns was most important for getting A2C/A3C to converge on
Doom MyWayHome.

\begin{table}[h]
\caption{
  Hyperparameters for A2C.
}
\label{tab:a2c}
\vskip 0.15in
\begin{center}
\begin{small}
\begin{sc}
\begin{tabular}{lr}
\toprule
Hyperparameter          & Doom \\
\midrule
Adam learning rate      & $1\mathrm{e}^{-4}$ \\
Adam minibatch size     & $640$ \\
entropy bonus ($\beta$) & $0.01$ \\
gradient clip           & $0.5$ \\
$n$-steps               & $40$ \\
num.~workers            & $16$ \\
\bottomrule
\end{tabular}
\end{sc}
\end{small}
\end{center}
\vskip -0.1in
\end{table}

\subsubsection{DQN}

Please see Table \ref{tab:dqn} for hyperparameters used with deep Q-learning.
Dueling double DQN uses the tuned hyperparameters
(van Hasselt et al., 2016; Wang et al., 2016).
In particular, we found that dueling double DQN generally performed better
and was more stable when learning on Atari with the tuned learning rate
$6.25\times10^{-5} \approx 6\times10^{-5}$
from Wang et al. (2016),
compared to the slightly larger learning rate of $1\times10^{-4}$
used by ARM.

\begin{table}[h]
\caption{
  Hyperparameters for dueling + double deep Q-learning.
}
\label{tab:dqn}
\vskip 0.15in
\begin{center}
\begin{small}
\begin{sc}
\begin{tabular}{lrrr}
\toprule
Hyperparameter          & Atari                 & Doom                  & Minecraft \\
\midrule
Adam learning rate      & $6\mathrm{e}^{-5}$    & $1\mathrm{e}^{-5}$    & $1\mathrm{e}^{-5}$ \\
Adam minibatch size     & $32$                  & $32$                  & $32$ \\
final exploration       & $0.01$                & $0.01$                & $0.01$ \\
gradient clip           & $10$                  & ---                   & --- \\
$n$-steps               & $1$                   & $5$                   & $5$ \\
replay memory init      & $50000$               & $50000$               & $12500$ \\
replay memory max       & $10^6$                & $240000$              & $62500$ \\
sim steps/grad step     & $4$                   & $4$                   & $4$ \\
target update steps     & $30000$               & $30000$               & $12500$ \\
\bottomrule
\end{tabular}
\end{sc}
\end{small}
\end{center}
\vskip -0.1in
\end{table}

\subsubsection{TRPO}

Please see Table \ref{tab:trpo} for hyperparameters used with TRPO.
We generally used the defaults, such as the KL step size of $0.01$
which we found to be a good default.
Decreasing the batch size improved sample efficiency on Doom and Minecraft
without adversely affecting the performance of the learned policies.

\begin{table}[h]
\caption{
  Hyperparameters for TRPO.
}
\label{tab:trpo}
\vskip 0.15in
\begin{center}
\begin{small}
\begin{sc}
\begin{tabular}{lrrr}
\toprule
Hyperparameter          & Atari                 & Doom                  & Minecraft \\
\midrule
batch size              & $100000$              & $12500$               & $6250$ \\
CG dampening            & $0.1$                 & $0.1$                 & $0.1$ \\
CG iterations           & $10$                  & $10$                  & $10$ \\
KL step size            & $0.01$                & $0.01$                & $0.01$ \\
\bottomrule
\end{tabular}
\end{sc}
\end{small}
\end{center}
\vskip -0.1in
\end{table}

\subsection{Environment and Task Details}

Our task-specific implementation details are described below.

\subsubsection{Atari}

For the occluded variant of Pong,
we set the middle region of the $160\times210$ screen
with $x,y$ pixel coordinates $[55 \ldots 105),[34 \ldots 194)$
to the RGB color $(144,72,17)$.
The image of occluded Pong in Figure 4 from the main text has a slightly
darker occluded region for emphasis.

We use the preprocessing and convolutional network model of Mnih et al.~(2013).
Specifically, we view every 4th emulator frame,
convert the raw frames to grayscale,
and perform downsampling to generate a single observed frame.
The input observation of the convnet is a concatenation of the most recent frames
(either 4 frames or 1 frame).
The convnet consists of an $8 \times 8$ convolution with stride 4 and 16 filters
followed by ReLU,
a $4 \times 4$ convolution with stride 2 and 32 filters followed by ReLU,
a linear map with 256 units followed by ReLU,
and a linear map with $|\mathcal A|$ units
where $|\mathcal A|$ is the action space cardinality
($|\mathcal A| = 6$ for Pong).

\subsubsection{Doom}

Our modified environment ``Doom Corridor+'' is very closely derived from the
default ``Doom Corridor'' environment in ViZDoom.
We primarily make two modifications:
(a) first, we restrict the action space to the three keys
$\{MoveRight,TurnLeft,TurnRight\}$,
for a total of $2^3=8$ discrete actions;
(b) second, we set the difficulty (``Doom skill'') to the maximum of 5.

For the occluded variant of Corridor+,
we set the middle region of the $160\times120$ screen
with $x,y$ pixel coordinates $[30 \ldots 130),[10 \ldots 110)$
to black, i.e.~$(0,0,0)$.

For Corridor+, we scaled rewards by a factor of $0.01$.
We did not scale rewards for MyWayHome.

The Doom screen was rendered at a resolution of $160 \times 120$ and downsized
to $84 \times 84$.
Only every 4th frame was rendered,
and the input observation to the convnet is a concatenation of
the last 4 rendered RGB frames for a total of 12 input channels.
The convnet contains 3 convolutions with 32 filters each:
the first is size $8 \times 8$ with stride 4,
the second is size $4 \times 4$ with stride 2,
and the third is size $3 \times 3$ with stride 1.
The final convolution is followed by a linear map with 1024 units.
A second linear map yields the output.
Hidden activations are gated by ReLUs.

\subsubsection{Minecraft}

Our Minecraft tasks are based on the tasks introduced by
Matiisen et al.~(2017), with a few differences.
Instead of using a continuous action space,
we used a discrete action space with 4 move and turn actions.
To aid learning on the last level (``L5''),
we removed the reward penalty upon episode timeout
and we increased the timeout on ``L5''
from 45 seconds to 75 seconds
due to the larger size of the environment.
We scaled rewards for all levels by $0.001$.

We use the same convolutional network architecture for Minecraft as we
use for ViZDoom.
The Minecraft screen was rendered at a resolution of $320 \times 240$ and
downsized to $84 \times 84$.
Only every 5th frame was rendered,
and the input observation of the convnet is a concatenation of
the last 4 rendered RGB frames for a total of 12 input channels.

\section{Off-policy ARM via Importance Sampling}

Our current approach to running ARM with off-policy data consists of applying
an importance sampling correction directly to the $n$-step returns.
Given the behavior policy $\mu$ under which the data was sampled,
the current policy $\pi_t$ under which we want to perform estimation,
and an importance sampling weight clip $c$ for variance reduction,
the corrected $n$-step return we use is:
\begin{align}
  g_k^n(\mu \| \pi_t)
  &= \sum_{k'=k}^{k+n-1} \gamma^{k'-k} \left( \prod_{\ell=k}^{k'} w_{\mu \| \pi_t}(a_\ell|o_\ell) \right) r_{k'} \\
  &~~~~~ + \gamma^n V'(o_{k+n};\varphi) \nonumber
\end{align}
where the truncated importance weight $w_{\mu \| \pi_t}(a|o)$ is defined:
\begin{align}
  w_{\mu \| \pi_t}(a|o)
  &= \min\left( c , \frac{\pi_t(a|o)}{\mu(a|o)} \right).
\end{align}
Note that the target value function $V'(o_{k+n};\varphi)$ does not require an
importance sampling correction because $V'$ already approximates the
on-policy value function $V_{\pi_t}(o_{k+n};\theta_t)$.
Our choice of $c=1$ in our experiments was inspired by Wang et al.,~(2017).
We found that $c=1$ worked well but note other choices for $c$ may also be
reasonable.

When applying our importance sampling correction,
we preserve all details of the ARM algorithm except for two aspects:
the transition sampling strategy (a finite memory of previous batches are
cached and uniformly sampled)
and the regression targets for learning the value functions.
Specifically,
the regression targets $v_k$ and $q_k^+$
(Equations (11)--(14) in the main text)
are modified to the following:
\begin{align}
  v_k
  &= g_k^n(\mu \| \pi_t) \\
  q_k
  &= (1 - w_{\mu \| \pi_t}(a_k|o_k)) r_k + g_{k}^{n}(\mu \| \pi_t) \\
  \phi_k
  &= Q^+_{t-1}(o_k,a_k;\omega_{t-1}) - V_{\pi_{t-1}}(o_k;\theta_{t-1}) \\
  q^+_k
  &= \max( 0 , \phi_k ) + q_k.
\end{align}

\section{Additional Experiments}

\subsection{Recurrence in Doom MyWayHome}

We evaluated the effect of recurrent policy and value function estimation
in the maze-like MyWayHome scenario of ViZDoom.
For the recurrent policy and value function, we replaced the first
fully connected operation with an LSTM featuring an equivalent number of
hidden units (1024).
We found that recurrence has a small positive effect on the convergence of
A2C,
but was much less significant than the choice of algorithm;
compare Figure 2~in the main text with Figure \ref{fig:doom_lstm} below.

\begin{figure}[h]
  \centering
  \includegraphics[width=0.5\linewidth]{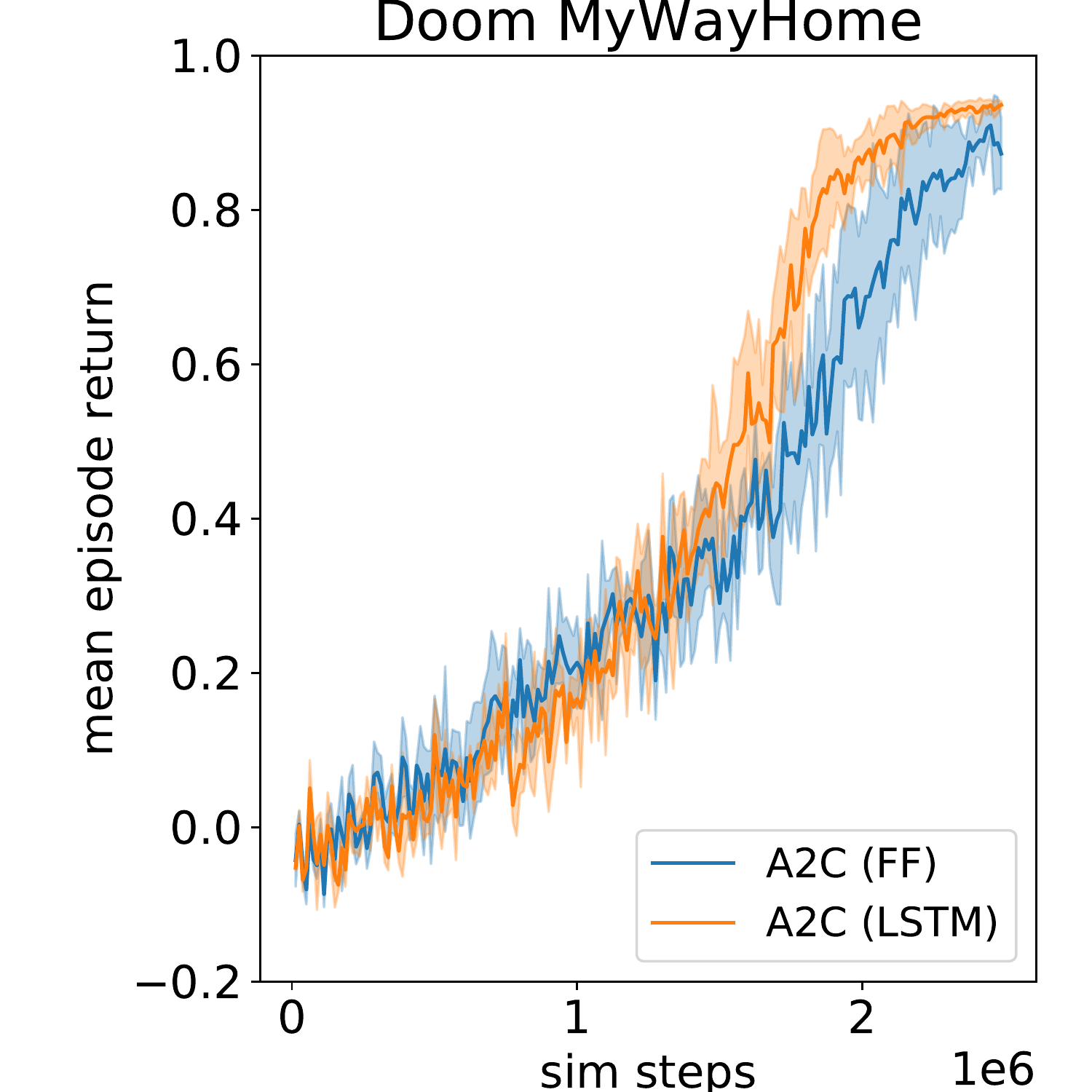}
  \caption{
    Comparing A2C with a feedforward convolutional network (blue)
    and a recurrent convolutional-LSTM network (orange)
    on the ViZDoom scenario MyWayHome.
  }
  \label{fig:doom_lstm}
\end{figure}

\subsection{Atari 2600 games}

\begin{figure*}[ht]
  \centering
  \includegraphics[width=0.20\linewidth]{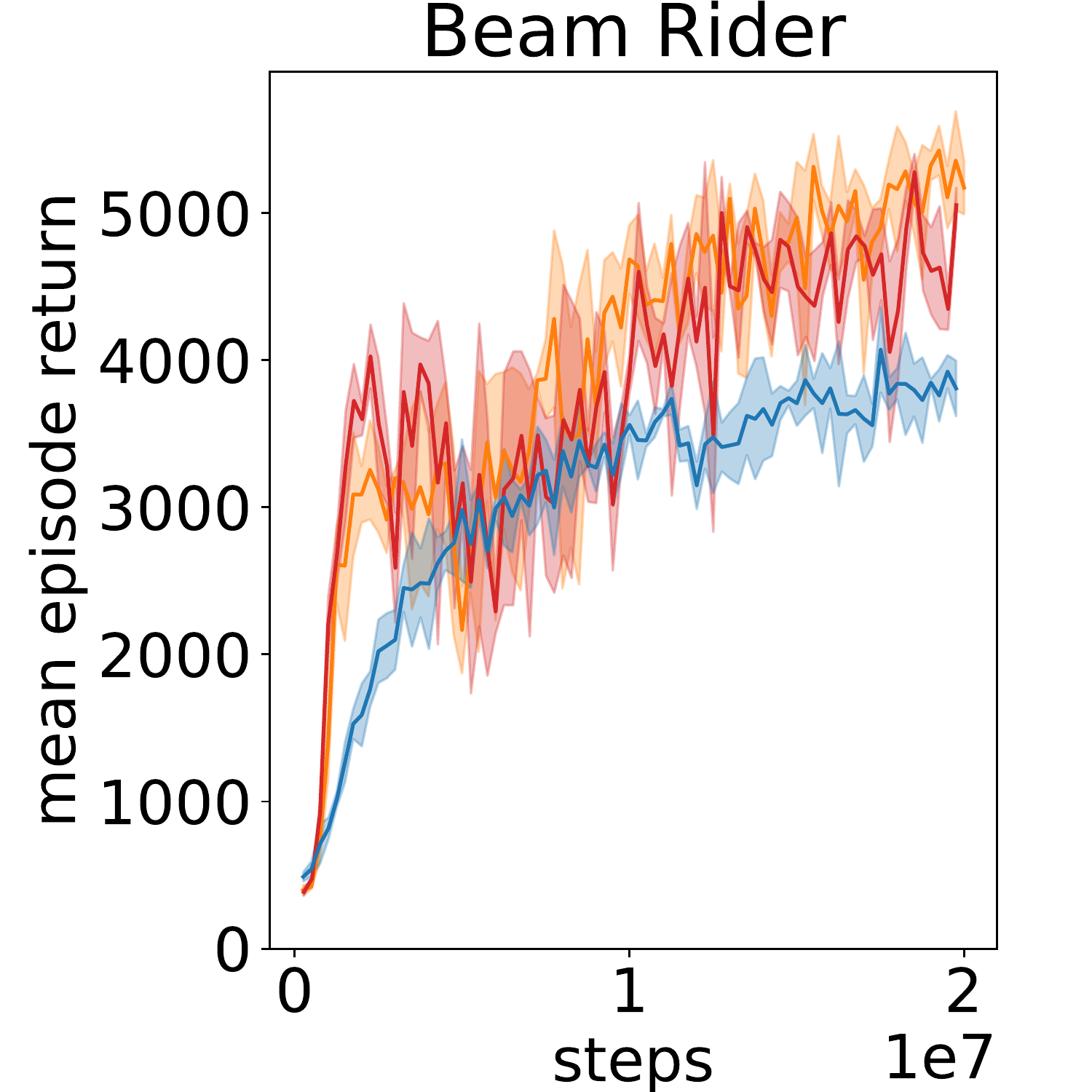}
  \includegraphics[width=0.20\linewidth]{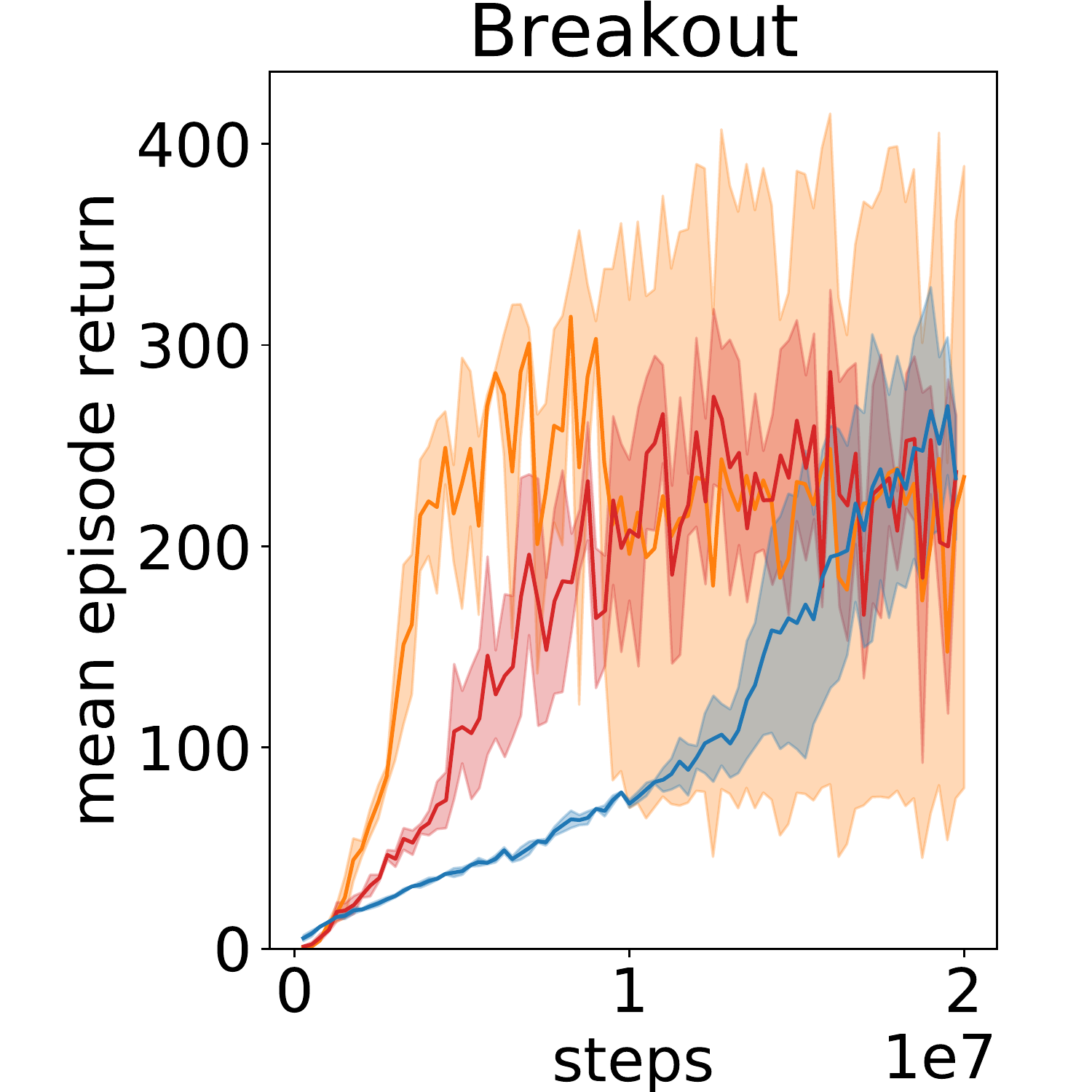}
  \includegraphics[width=0.20\linewidth]{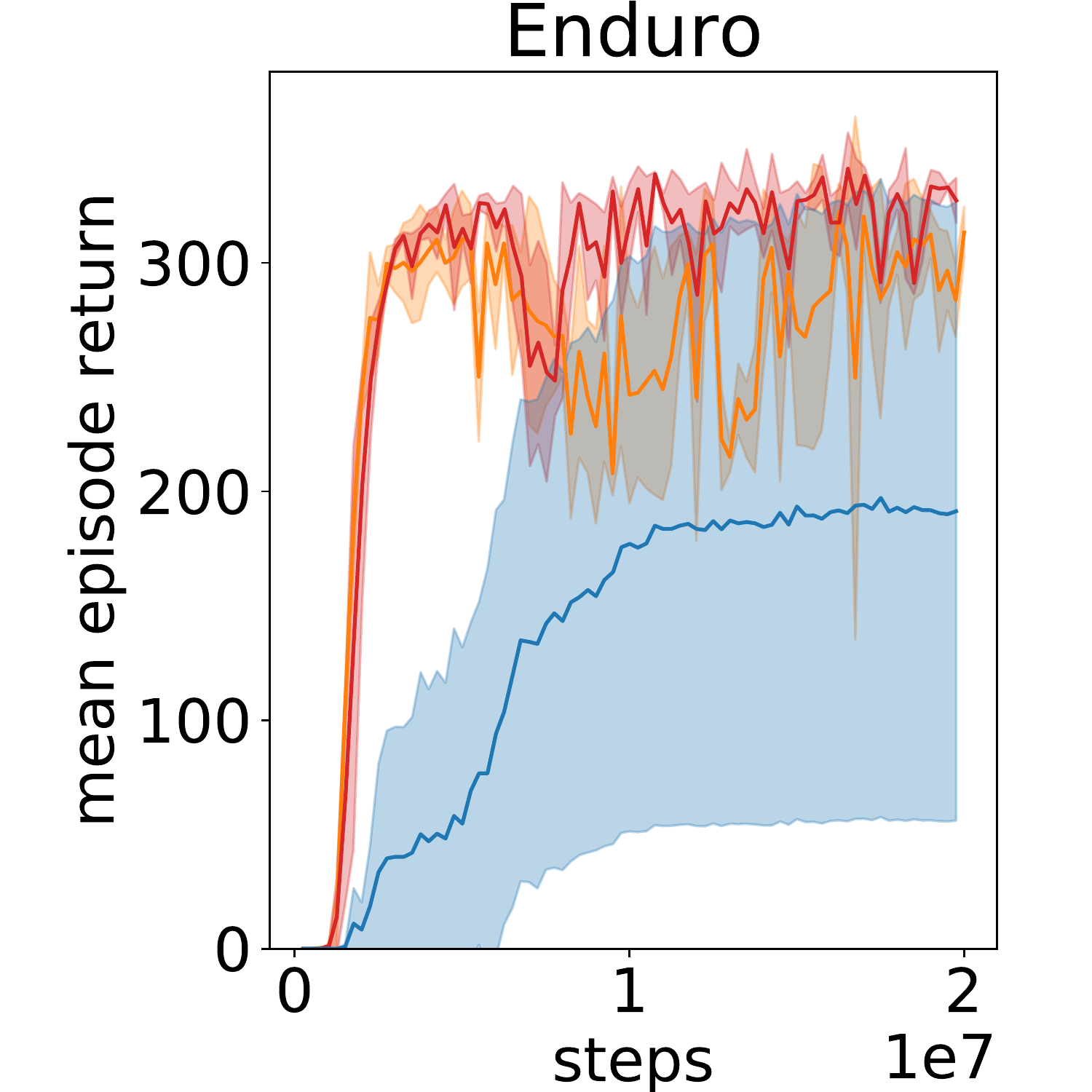}
  \includegraphics[width=0.20\linewidth]{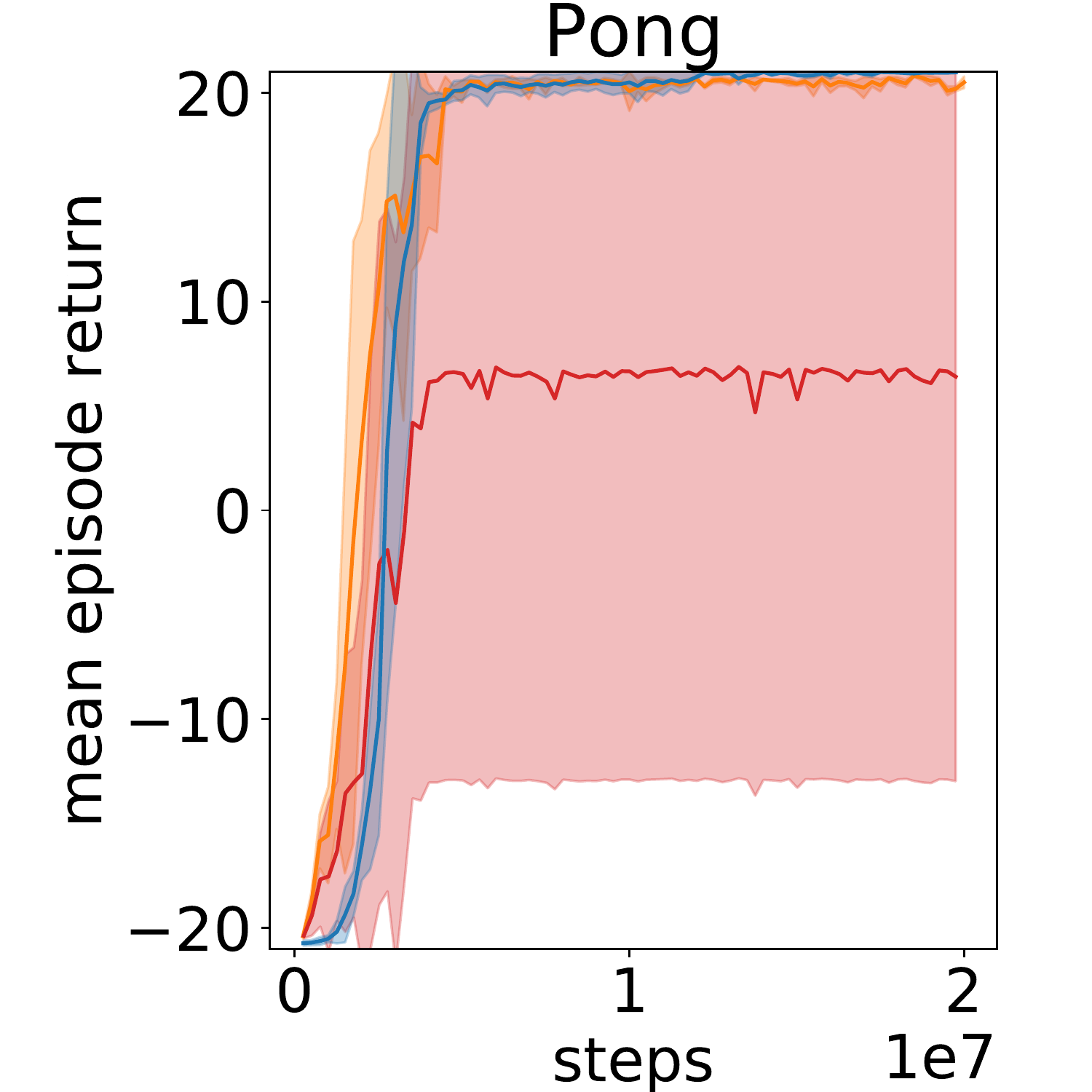} \\
  \includegraphics[width=0.20\linewidth]{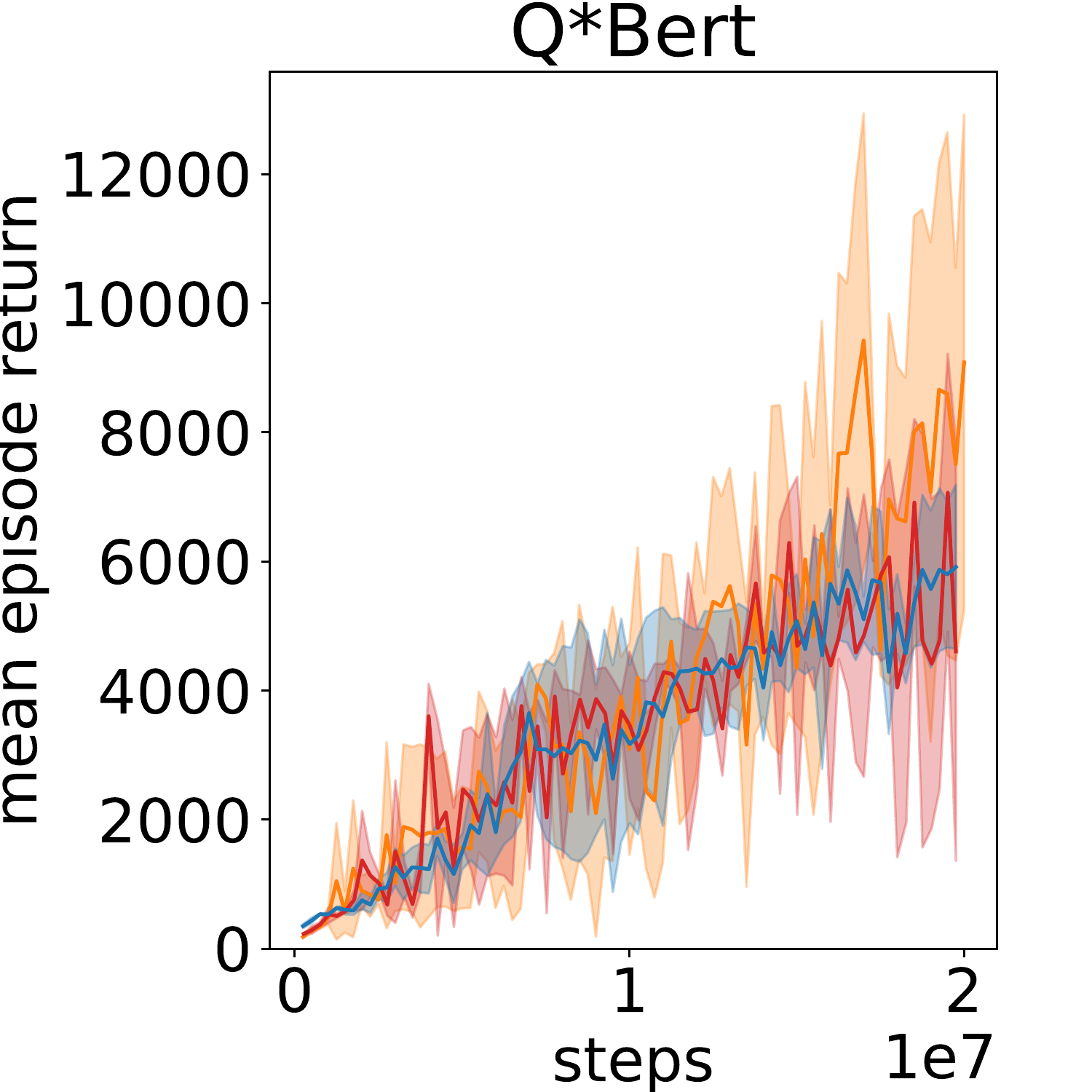}
  \includegraphics[width=0.20\linewidth]{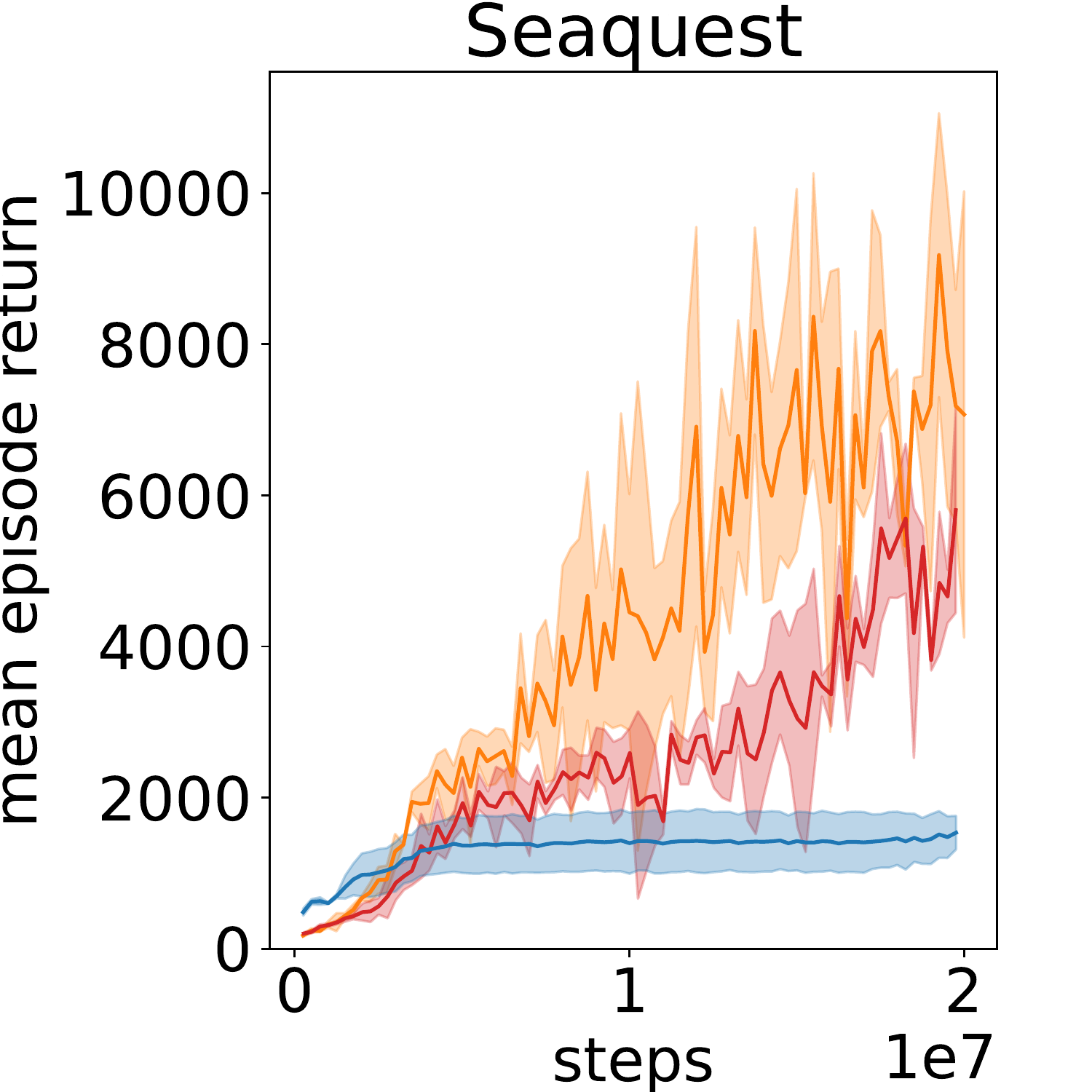}
  \includegraphics[width=0.20\linewidth]{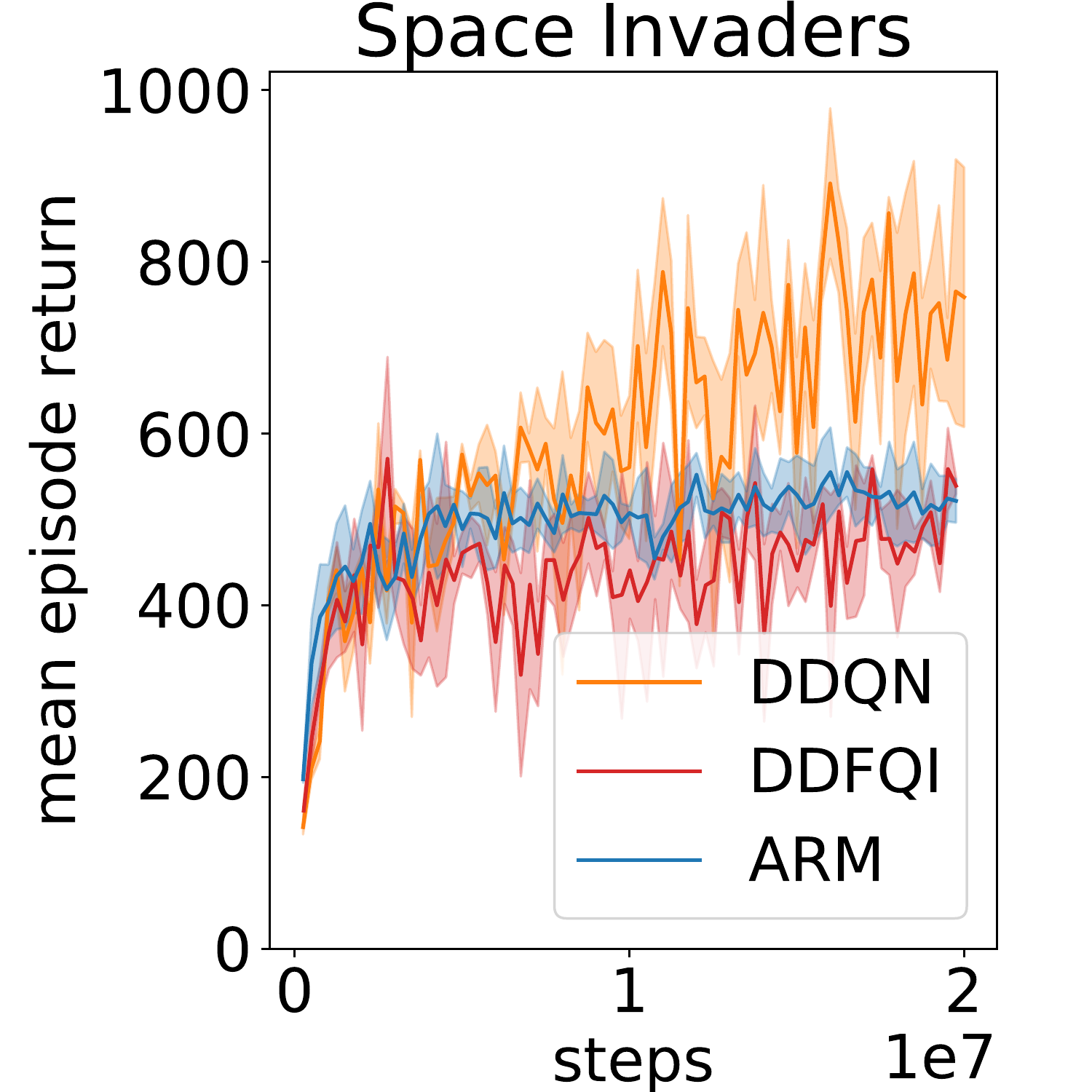}
  \caption{
    Comparing double deep Q-learning (orange),
    double deep fitted Q-iteration (red),
    and ARM (blue)
    on a suite of seven Atari games from the Arcade Learning Environment.
  }
  \label{fig:atari}
\end{figure*}

Although our primary interest is in partially observable reinforcement learning
domains,
we also want to check that ARM works in nearly fully observable and Markovian
environments,
such as Atari 2600 games.
We consider two baselines: double deep Q-learning,
and double deep fitted Q-iteration which is a batch counterpart to double DQN.
We find that double deep Q-learning is a strong baseline for learning to play
Atari games,
although ARM still successfully learns interesting policies.
One major benefit of Q-learning-based methods is the ability to utilize a
large off-policy replay memory.
Our results on a suite of Atari games are in Figure \ref{fig:atari}.

\end{document}